\documentclass[journal,10pt]{IEEEtran}
\usepackage{amsthm}
\usepackage{amsfonts}
\usepackage{amssymb}
\usepackage{mathrsfs}
\usepackage{mathtools}
\usepackage{float}
\usepackage[pdftex]{graphicx}
\usepackage{amsmath}
\usepackage{color}
\usepackage{multirow}
\usepackage{graphicx}
\usepackage{graphicx,float,wrapfig,epstopdf,amsmath}
\usepackage{comment}
\usepackage{balance}
\usepackage{lipsum}
\usepackage[inline]{enumitem}
\usepackage{cuted}
\usepackage[caption=false,font=footnotesize]{subfig}
\usepackage{cite}
\usepackage{hyperref}
\usepackage{enumitem}
\usepackage[]{algorithmicx}
\usepackage{algpseudocode,algorithm}

\newlist{myenumi}{description}{10}
\setlist[myenumi]{labelindent=\parindent, leftmargin=*, label=(\roman*), align=left}
\setlist[myenumi]{leftmargin=0pt}
\newtheorem{theo}{Theorem}
\newtheorem{lem}{Lemma}

\theoremstyle{definition}

\hypersetup{%
	colorlinks=true,
	linkcolor={red},
	raiselinks=false
}
\DeclareMathOperator*{\subjectto}{subject \hspace{3pt} to:\hspace{3pt}} 
\usepackage[table]{xcolor}
\definecolor{color1}{RGB}{255,255,255}
\definecolor{color2}{RGB}{255,255,255}
\usepackage{slashbox}

\begin{document}


	
	\title{A Hybrid Architecture for  \\ Federated and Centralized Learning }

	\author{\IEEEauthorblockN{Ahmet M. Elbir, \textit{Senior Member IEEE}, Sinem Coleri, \textit{Fellow IEEE}, Anastasios K. Papazafeiropoulos, \textit{Senior Member IEEE}, Pandelis Kourtessis and  Symeon Chatzinotas, \textit{Senior Member IEEE}} 
		\thanks{This work was supported in part by the  ERC project  AGNOSTIC, and by the Scientific and Technological Research Council of Turkey with European CHIST-ERA grant 119E350.}
		\thanks{A preliminary work of this paper was presented in 2021 European Signal Processing Conference (EUSIPCO)~\cite{elbir2021HFCL}.}
		\thanks{A. M. Elbir is with Duzce University, Duzce, Turkey, and with SnT, University of Luxembourg, Luxembourg (e-mail: ahmetmelbir@gmail.com).} 
		\thanks{S. Coleri is with the Department of Electrical and Electronics Engineering, 
			Koc University, Istanbul, Turkey (e-mail: scoleri@ku.edu.tr).}
		\thanks{A. K. Papazafeiropoulos is with the CIS Research Group, University of Hertfordshire, Hatfield, U. K. and with SnT at the University of Luxembourg, Luxembourg. (e-mail: tapapazaf@gmail.com). }
		
		\thanks{P. Kourtessis is with the CIS Research Group, University of Hertfordshire, Hatfield, U. K. E-mail: p.kourtessis@herts.ac.uk}
		\thanks{S. Chatzinotas is with the SnT at the University of Luxembourg, Luxembourg. (e-mail:symeon.chatzinotas@uni.lu). }	
	}
	
	\maketitle

	\begin{abstract}
	Many of the machine learning tasks rely on centralized learning (CL), which requires the transmission of local datasets from the clients to a parameter server (PS) entailing huge communication overhead. To overcome this, federated learning (FL) has been suggested as a promising tool, wherein the clients send only the model updates to the PS instead of the whole dataset.  However, FL demands powerful computational resources from the clients. In practice, not all the clients have sufficient computational resources to participate in training. To address this common scenario, we propose a more efficient approach called hybrid federated and centralized learning (HFCL), wherein only the clients with sufficient resources employ FL, while the remaining ones send their datasets to the PS, which computes the model on behalf of them. Then, the model parameters are aggregated at the PS. To improve the efficiency of dataset transmission, we propose two different techniques: i) increased computation-per-client and ii) sequential data transmission. Notably, the HFCL frameworks outperform FL with up to $20\%$ improvement in the learning accuracy when only half of the clients perform FL while having $50\%$ less communication overhead than CL since all the clients collaborate on the learning process with their datasets.
		
	\end{abstract}

	\section{Introduction}
	The ever growing increase in the number of connected devices in the last few years has led to a surge in the amount of data generated by mobile phones, connected vehicles, drones and internet of things (IoT) devices due to the rapid development of various emerging applications, such as artificial intelligence (AI), virtual and augmented reality (VAR), autonomous vehicles (AVs), and machine-to-machine communications~\cite{deepLearningScience,survey_DL_Scalable}.  According to  international telecommunication union (ITU), the global mobile traffic is expected to reach $607$ EB by 2025. Moreover, AVs are expected to generate approximately 20 TB/day/vehicle~\cite{ML_for_VANET}. In order to process and extract useful information from the huge amount of data, machine learning (ML) has been recognized as a promising tool for  emerging technologies, such as IoT~\cite{ml4IoTSurvey}, AV~\cite{elbir2020federated}, and  next generation wireless communications~\cite{elbir2020cognitive,elbir2020DL4IRS_survey,mimoDLintro} due to its success in image/speech recognition, natural language processing, etc~\cite{deepLearningScience}. These applications require huge amount of data to be processed and learned by a learning model, often an artificial neural network (ANN), by extracting the features from the raw data and providing a ``meaning'' to the input via constructing a model-free data mapping with huge number of learnable parameters~\cite{elbir2020cognitive}. The implementation of these learning models demands powerful computation resources, such as graphics processing units (GPUs). Therefore, huge learning models, massive amount of training data, and  powerful computation infrastructure are the  main driving factors defining the success of ML algorithms~\cite{elbir2020cognitive,survey_DL_Scalable}.
	
	Many of the ML tasks are based on centralized learning (CL) algorithms, which train powerful ANNs at a parameter server (PS)~\cite{deepLearningScience,edgeWireless, fl_spm_federatedLearning}. While CL inherently assumes 	the availability of datasets at the PS, this may not be possible for the 	wireless edge devices (clients), such as mobile phones, connected vehicles, and IoT devices, which need to send their datasets to the PS. Transmitting the collected datasets to the PS in a reliable manner may be too costly in terms of	energy, latency, and bandwidth~\cite{FL_Gunduz}. For example, in  LTE (long term evolution), a single frame of $5$ MHz bandwidth and $10$ ms duration can carry only $6000$ complex symbols~\cite{FL_gunduz_fading}, whereas the size of the whole dataset can be on the order of hundreds of thousands symbols~\cite{elbirQuantizedCNN2019}. As a result, CL-based techniques require huge bandwidth  and communication overhead during training.

	\begin{figure*}[t]
		\centering
		\subfloat[] {\includegraphics[draft=false,scale=.09]{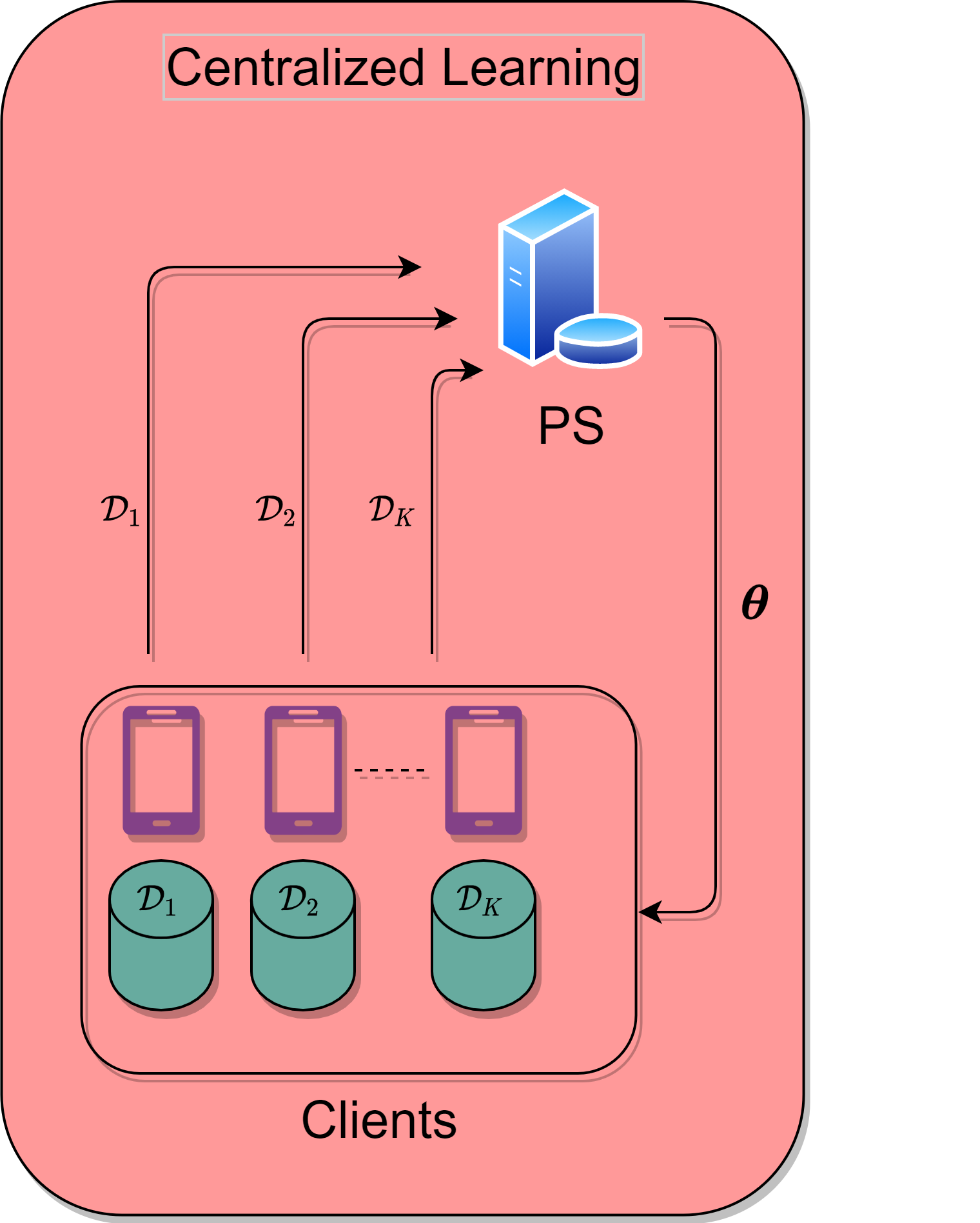}	\label{fig_Diaga} } 
		\subfloat[] {\includegraphics[draft=false,scale=.09]{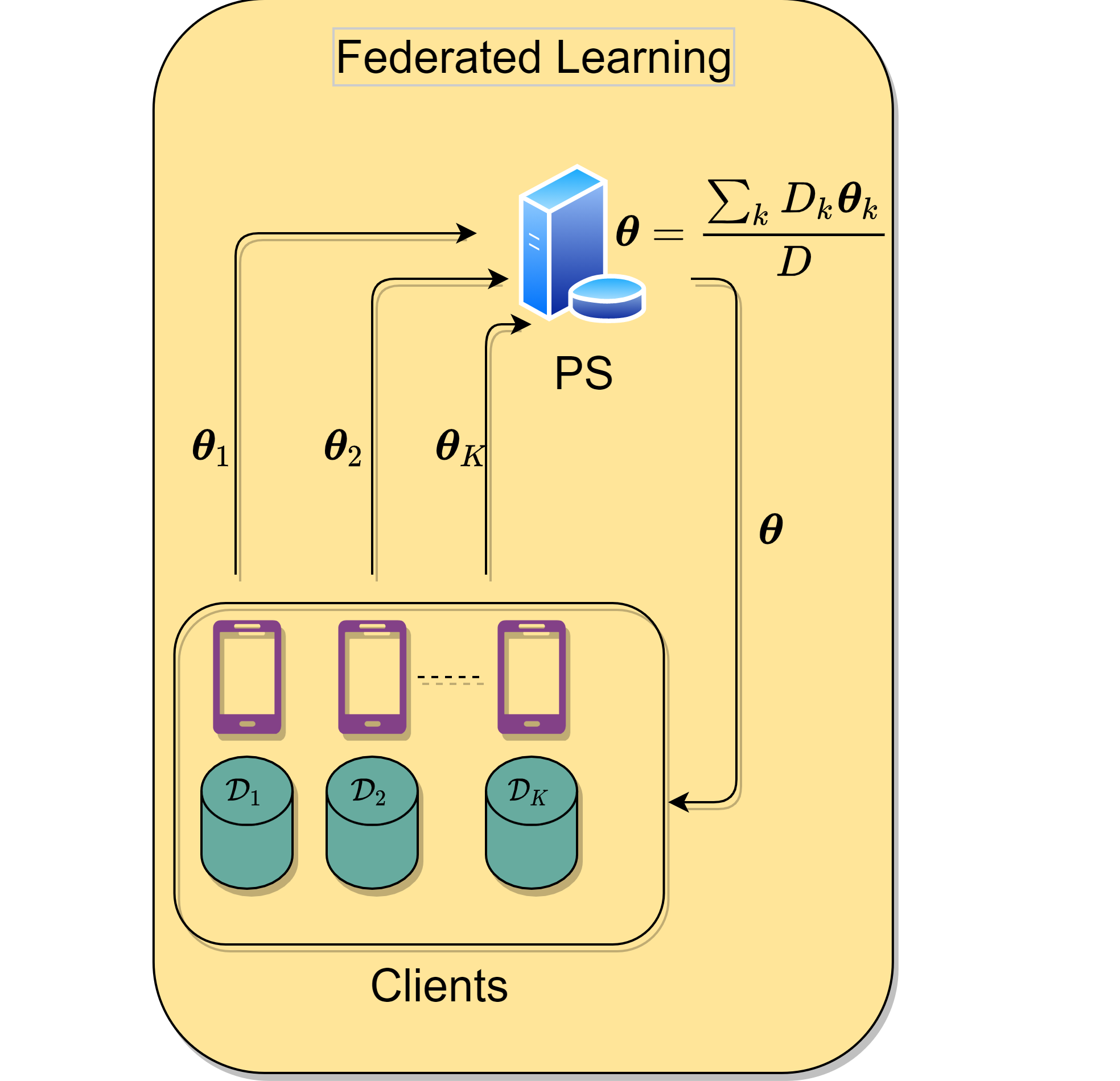} 	\label{fig_Diagb}} 
		\subfloat[] {\includegraphics[draft=false,scale=.09]{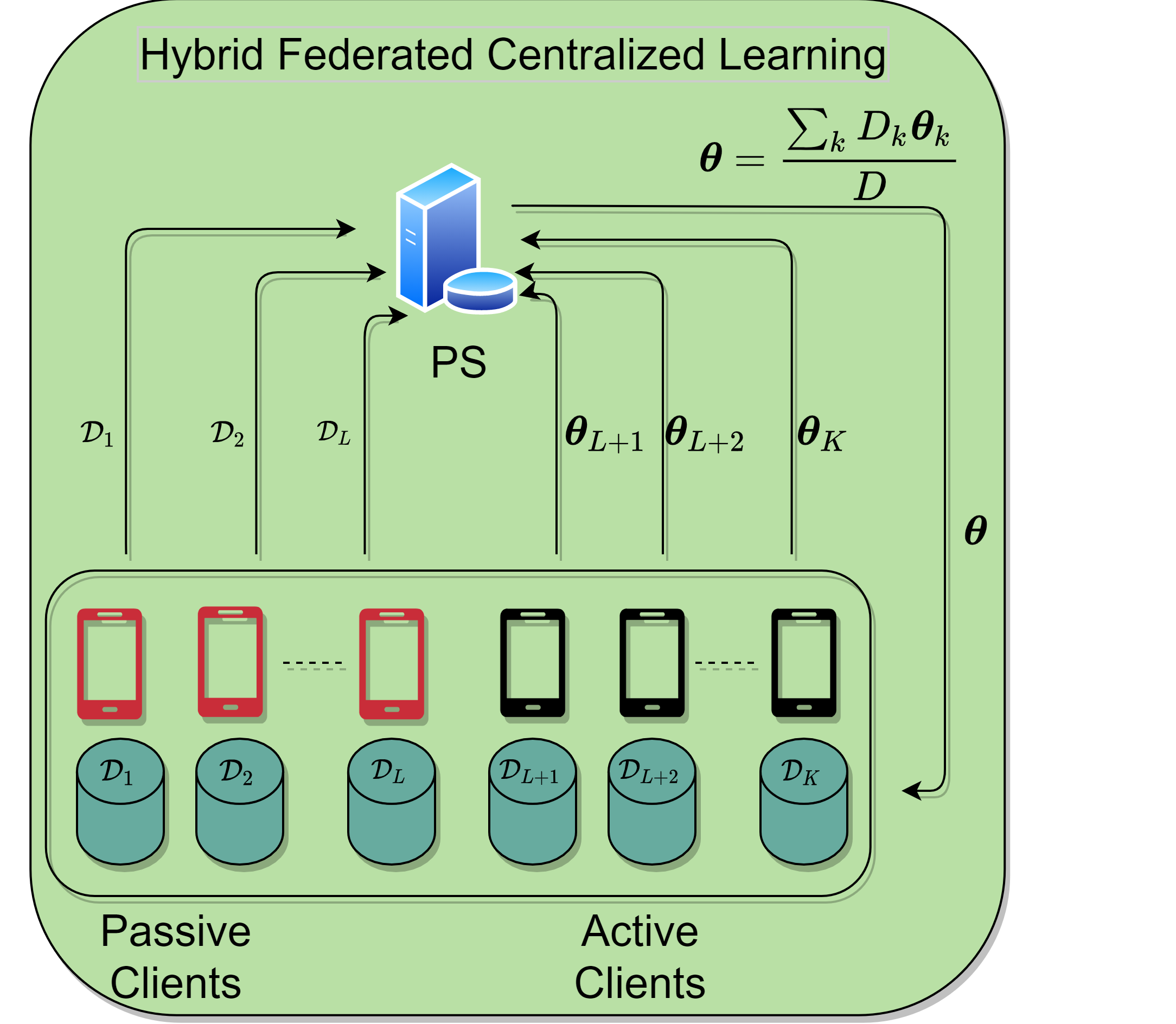} 	\label{fig_Diagc}} 
		\caption{CL, FL and HFCL frameworks. (a) In CL, all clients transmit their datasets to the PS.  (b) In FL, the datasets are preserved at the clients while model parameters are sent to the PS. (c) In HFCL, the clients are designated as active and inactive depending on their computational capability to either perform CL or FL.
		}
		\label{fig_Diag}
	\end{figure*}

	In order to provide a practically viable alternative to CL-based training, federated learning (FL) has been proposed to exploit the processing capability of the edge devices and the local datasets of the clients~\cite{fl_spm_federatedLearning,fl_By_Google,fl_applications}. In FL, the clients compute and transmit the model parameters to the PS instead of their local datasets as in CL to collaboratively train the learning model. The collected model updates are aggregated at the PS and then broadcast to the clients to further update the learning parameters iteratively. \textcolor{black}{Since FL does not access the whole dataset at once, it has slightly lower learning performance than that of CL. However, FL is communication-efficient and privacy-preserving since it keeps the datasets at the clients.} Recently, FL has been applied to image classification~\cite{fl_By_Google,FL_Gunduz,fl_convergenceOnNIIDData}, speech recognition~\cite{fl_speechRecognition} and wireless communications~\cite{elbir2020FL_HB,elbir2020_FL_CE}. In particular, various wireless network architectures exploiting FL have been investigated, such as cellular networks~\cite{irs_FL_BF_fromChannel,FL_Bennis3,FL_gunduz_fading}, vehicular networks~\cite{elbir2020federated}, unmanned aerial vehicles~\cite{FL_Bennis2},  IoT networks~\cite{fl_IoT}, and for the design of physical layer  applications~\cite{elbir2021FL_PHY}. In these works, the FL architectures rely on the fact that all of the clients are capable of model computation, which may require powerful parallel processing units, such as GPUs. However, this may not always be possible in practice due to the diversity of the devices with different computation capabilities, such as mobile phones, vehicular components and IoT devices. When the edge devices lack of sufficient  computational power, they cannot perform model computation, and thus become unable to participate in the learning process. To address this problem, client selection algorithms have been developed. In~\cite{FL_vehicular1} and \cite{fl_clientSelection}, trusted clients and the ones with sufficient computational resources are participated in FL-based training. However, these studies do not allow the clients that are not selected to participate in the training process, regardless of their computational capability.

	In this work, we introduce a hybrid FL and CL (HFCL) framework {\color{black} for which the main motivation is to effectively train a learning model regardless of the computational capability of the clients}. By exploiting their computational resources, the clients that are capable of model computation perform FL while the remaining clients resort to CL and send their datasets to the PS, which computes the corresponding model parameters on behalf of them (see Fig.~\ref{fig_Diag}).  At the beginning of the training, the clients are designated as \emph{inactive} (i.e.,  CL is applied) or \emph{active} (i.e.,  FL is employed) depending on their computational capabilities. \textcolor{black}{That is, the clients are called active if they have adequate computational resources, e.g., GPUs, to compute the model parameters, whereas the clients are called inactive otherwise in such a case, hence they send their datasets to the PS, which computes the model parameters on behalf of them. The PS, then, optimizes the bandwidth allocation by minimizing the maximum communication delay of the clients.  During model training, both active and inactive clients are synchronized and} the training process is conducted iteratively as in FL, where the model parameters are aggregated in each communication round between the PS and the active clients. 	Prior to the model training, the inactive clients transmit the datasets to the PS so that the PS can compute model updates  on behalf of them during training. In the meantime, the active clients perform model computation. Once the model parameters corresponding to active (computed on device) and inactive (computed on PS) clients are collected at the PS, they are aggregated and broadcast to the active clients only so that they can compute the model parameters in the next iteration. Since the PS computes the model parameters on behalf of the inactive clients, it is not necessary to broadcast model updates to them. As a result, the proposed HFCL approach provides a trade-off among learning performance, communication overhead, privacy, and the clients' computational resources (see, e.g., Table~\ref{tableComparison}). While the usage of CL enables higher learning performance and lower dependence on the clients' computational resources, FL enjoys enhanced privacy and less communication overhead. In contrast, HFCL jointly employs CL and FL. Therefore, it enjoys less communication overhead than CL; higher performance and more flexibility than FL, while it is less privacy-preserving and lower communication-efficient than FL. {\color{black}Thus, HFCL can be particularly useful when the dataset does not include privacy-sensitive content, e.g., physical layer data~\cite{elbir2021FL_PHY}.  }
	
	{\color{black}
	Although the integration of FL and CL may seem straightforward, there are two main challenges in HFCL. First, training time of HFCL is longer than FL since the inactive clients need to transmit their datasets to the PS prior to the training. In other words, the active clients should wait for the inactive clients at the beginning of the training until the dataset transmission for inactive clients is completed so that they can synchronously participate in model training. To circumvent this problem and provide efficient model training, we propose sequential dataset transmission (HFCL-SDT) where the inactive clients' datasets are divided into smaller blocks to complete  the dataset transmission quicker. Second, the learning accuracy of HFCL is lower than CL since the model parameters corresponding to the active clients are noisy due to wireless transmission. To improve the learning accuracy, we propose increased computation-per-client (HFCL-ICpC) approach where the	training continues at the active clients during the transmission of inactive clients' datasets instead of waiting idle for the inactive 	clients as in basic HFCL, in which the active clients wait for the completion of dataset transmission. Thus, HFCL-ICpC improves the learning performance as compared to HFCL without increasing the delay due to dataset generation.} In summary, HFCL-ICpC improves the learning performance while HFCL-SDT reduces the dataset transmission time per communication round. \textcolor{black}{Inspired from~\cite{fl_By_Google}, where the model parameters are infrequently aggregated at the clients, we develop a new approach to first aggregate the model parameters at the clients until the transmission of the inactive clients' datasets is completed, then continue aggregating the models at the PS. Compared to~\cite{fl_By_Google}, our technique achieves a  better learning performance as well as flexibility on the hardware requirements since the learning model can be trained on the whole dataset even if a part of the clients do not have computational resources.} Compared to our preliminary work~\cite{elbir2021HFCL}, this paper introduces HFCL-ICpC approach to improve learning accuracy, theoretical analysis on the convergence of HFCL, and the implementation of the 3-D object detection scenario. {\color{black}While the term \emph{hybrid} is used in some recent FL-based works in terms of client scheduling~\cite{Shi2020Dec}, data partitioning~\cite{Zhang2020Dec} and data security~\cite{Huang2021Aug},  this is the first work employing a hybrid architecture of FL and CL by exploiting the hardware capability of the edge devices.} The main contributions of this paper are as follows:
	\begin{itemize}
		\item We propose a hybrid FL/CL approach to manage the diverse  computational capabilities of the edge clients. In the proposed approach, only the clients, which have adequate computational resources perform FL while the remaining clients perform CL. 
		\item \textcolor{black}{We show that the performance of HFCL converges to that of FL (CL) when the number of active clients increases (decreases). Specifically when the active and inactive clients are partitioned equally, the proposed HFCL approach exhibits up to $20\%$  improvement in the learning accuracy compared to FL. HFCL also enjoys $50\%$ less communication overhead with only $2\%$ loss in the learning accuracy as compared to CL. }
		\item To efficiently transmit the datasets of the inactive clients to the PS, we propose the HFCL-SDT and HFCL-ICpC techniques to achieve no interruption of the learning process. Compared to FL, the proposed techniques are effective due to faster convergence rates and higher learning accuracy together with less communication delay due to less dataset transmission.

	\end{itemize}
	
	%

	\begin{table}[t]
		\caption{Comparison of CL, FL and HFCL}
		\label{tableComparison}
		\centering
		\begin{tabular}{c|c|c|c }
			\hline
			\hline
			\backslashbox{\textbf{Property}}{\bf Framework}\cellcolor{color1} &\bf CL  
			\cellcolor{color2}  
			&\bf FL
			\cellcolor{color1} &\bf HFCL\cellcolor{color2}  \\
			\textbf{Communication Overhead}\cellcolor{color2} & High\cellcolor{color1} & Low\cellcolor{color2} &Moderate\cellcolor{color1}  \\
			\hline
			\textbf{Learning Accuracy}\cellcolor{color1} & High\cellcolor{color2} & Moderate\cellcolor{color1} &Moderate\cellcolor{color2} \\
			\hline
			\hspace{-3pt}\textbf{Clients' Hardware Requirement}\hspace{-3pt}\cellcolor{color2} & \cellcolor{color1}Low& High\cellcolor{color2} &Flexible \cellcolor{color1} \\
			\hline
			\textbf{Privacy-Preserving}\cellcolor{color1} & Low\cellcolor{color2} & High\cellcolor{color1} &Moderate\cellcolor{color2} \\
			\hline
			\hline
		\end{tabular}
	\end{table}

	%

	\textit{Notation:} Throughout the paper, the identity matrix of size $N\times N$ is denoted by $\mathbf{I}_N$. The notation $(\cdot)^\textsf{T}$ denotes the transpose  operation. The notations  $[\mathbf{A}]_{i,j}$ and $[\mathbf{a}]_i$ denote the $(i,j)$th element of matrix $\mathbf{A}$ and the $i$th element of vector $\mathbf{a}$, respectively.  The function $\mathbb{E}\{\cdot\}$ expresses the statistical expectation of its argument while  $\|\mathbf{a}\|$ denotes the $l_2$-norm and $\nabla$ represents the gradient of vector quantity. A convolutional layer with $N$ $D\times D$ 2-D kernels is represented by $N$@ $D\times D$.
	
	The rest of the paper is organized as follows. In Section~\ref{sec:Preliminary}, CL and FL frameworks are presented. Then, we introduce the proposed HFCL, HFCL-ICpC, and HFCL-SDT approach in Section~\ref{sec:HCFL},~\ref{sec:HFCLicpc}, and~\ref{sec:HFCLsdt}.  In Section~\ref{sec:ConvergenceOverhead}, we investigate the convergence and the communication overhead of the HFCL approach. Section~\ref{sec:sim} presents the numerical simulation results, and we summarize the paper in Section~\ref{sec:conc} with concluding remarks.

	{
	\begin{table}[t]
		\color{black}
		\caption{Summary of Main Notations }
		\label{tableNotation}
		\centering
		\begin{tabular}{c|c }
			\hline
			\hline
		 \bf Notation &\bf Definition \\   
			 \hline
			 $\mathcal{D}$& The whole dataset  \\
			 \hline
			 $\mathcal{X}$& Input data \\
			 \hline
			 $\mathcal{Y}$& Output data  \\
			 \hline
			 $\boldsymbol{\theta}$& Model parameters  \\
			 \hline
			 $\mathcal{F}(\boldsymbol{\theta})$& Loss function (\ref{lossRegression}), (\ref{lossClassification})  \\
			 \hline
			 $\mathbf{g}(\boldsymbol{\theta})$ &Gradient vector (\ref{eq:update1}) \\ 
			 \hline
			 $\mathcal{L}$& Set of inactive clients  \\ 
			 \hline
			  $\bar{\mathcal{L}}$& Set of active clients  \\ 
			 \hline
			 $\tilde{ \boldsymbol{\theta}}$&Noisy model parameters (\ref{eq5})   \\ 
			 \hline
			$\tilde{\boldsymbol{\theta}}_k$ & Noisy model received at the $k$th active client  (\ref{noiseAtActiveClients})\\
			 \hline
			 $\bar{\mathcal{F}}_k(\boldsymbol{\theta})$& Regularized loss function for active clients (\ref{lossFunctionModified})  \\
			 \hline
			 $\tilde{\mathcal{F}}_k(\boldsymbol{\theta})$& Regularized loss function for inactive clients (\ref{lossPassive})   \\
			 \hline
			 $\eta$& Learning rate  \\
			 \hline
			$Q$ & Data block size during dataset transmission  \\
			 \hline
			 $N$& Number of local model updates   \\
			 \hline
			 $d_k$&Number of transmitted symbols   \\
			 \hline
			 $\tilde{\sigma}^2$& Noise variance of $\widetilde{\Delta\boldsymbol{\theta}}$ (\ref{noiseAtClient})  \\ 
			 \hline
			${\sigma}_k^2$& Noise variance of $\widetilde{\Delta\boldsymbol{\theta}}_k$ (\ref{noiseAtActiveClients})  \\   
			\hline
			\hline
		\end{tabular}
	\end{table}
}

	\section{Preliminaries: \\ Centralized and Federated Learning}
	\label{sec:Preliminary}
	In ML, we are interested in constructing a learning model that forms a non-linear relationship between the two data pairs:  the input and the label. Let $\mathcal{D}^{(i)} = (\mathcal{X}^{(i)},\mathcal{Y}^{(i)})$ be the $i$th tuple of the dataset $\mathcal{D}$ for $i = 1,2,\dots, D$, where $D = |\mathcal{D}|$ denotes the number of data instances in $\mathcal{D}$.  Here, $\mathcal{X}^{(i)}\in \mathbb{R}^{U_x\times V_x}$ and $\mathcal{Y}^{(i)}\in \mathbb{R}^{U_y\times V_y}$ denote the $i$th input and label pairs of $\mathcal{D}$, respectively. Eventually, the non-linear function, representing the ML model, can be given by $f(\mathcal{X}^{(i)}|\boldsymbol{\theta}) =\mathcal{Y}^{(i)}$, for which $\boldsymbol{\theta}\in \mathbb{R}^P$ denotes the vector of learnable model parameters. The training of the ML model takes place by focusing on a scenario, wherein
	$K$ clients collaborate on optimizing $\boldsymbol{\theta}$  for the ML task.
	
	\subsection{Centralized Learning} In CL, the model has access to the whole dataset $\mathcal{D}$, which is collected by the PS from the clients (see Fig.~\ref{fig_Diaga}). Let $\mathcal{D}_k$ be the local dataset of the $k$th client with $D_k$ being its size such that $\mathcal{D} = \bigcup_{k\in \mathcal{K}} \mathcal{D}_k$ and $D = \sum_{k \in \mathcal{K}} D_k$, where $\mathcal{K} = \{1,\dots,K\}$. In that event, {\color{black}the CL-based model training can be performed at the PS by solving the following optimization problem over the learnable parameters $\boldsymbol{\theta}$ as
	\begin{align}
	\label{lossML}
	\boldsymbol{\theta}^\star = \arg \min_{\boldsymbol{\theta}} \; &   
	\mathcal{F}(\boldsymbol{\theta}) =  \sum_{k = 1}^{K} \frac{D_k}{D}\mathcal{F}_k(\boldsymbol{\theta})
	, \nonumber\\
	\subjectto& f( \mathcal{X}_k^{(i)}|\boldsymbol{\theta}) = \mathcal{Y}_k^{(i)},\;\;\; i=1,\dots, D_k,
	\end{align}
	where $\boldsymbol{\theta}^\star$ denotes the model parameters after training.  $\mathcal{X}_k^{(i)}$ and $\mathcal{Y}_k^{(i)}$  denote respectively the input and output  of the $i$th element of $\mathcal{D}_k$ as $\mathcal{D}_k^{(i)} = (\mathcal{X}_k^{(i)},\mathcal{Y}_k^{(i)})$.  $\mathcal{F}_k(\boldsymbol{\theta})$ is the  loss function, and it can be defined for regression and classification tasks, respectively, as
	\begin{align}
	\label{lossRegression}
	\mathcal{F}_k(\boldsymbol{\theta}) =    \frac{1}{D_k} \sum_{i=1}^{D_k} \| f( \mathcal{X}_k^{(i)}|\boldsymbol{\theta}) - \mathcal{Y}_k^{(i)}  \|^2,
	\end{align}
	and 
	\begin{align}
	\label{lossClassification}
	\mathcal{F}_k(\boldsymbol{\theta})  =
	-\frac{1}{D_k} \sum_{i=1}^{D_k}& [\mathcal{Y}_k^{(i)} \ln f(\mathcal{X}_k^{(i)}|\boldsymbol{\theta})  \nonumber \\ & + (1 - \mathcal{Y}_k^{(i)}) \ln (1 - f(\mathcal{X}_k^{(i)}|\boldsymbol{\theta})) ].
	\end{align}}
	
	\subsection{Federated Learning}
	Compared to CL, FL does not involve the transmission of datasets to the PS. Instead, the model training is performed at the clients while the model parameters produced by  the clients are aggregated at the PS, as shown in Fig.~\ref{fig_Diagb}. {\color{black}Consequently, the solution of the optimization problem in FL settings takes place at the client and the PS side as follows
	\begin{subequations}
		\label{eq:updatePS_Client_1}
		\begin{align}
		\mathrm{Client:   }\hspace{10pt}& \boldsymbol{\theta}_k = \arg\min_{\boldsymbol{\theta}} \mathcal{F}_k(\boldsymbol{\theta}), \hspace{10pt} k \in \mathcal{K}, \label{FL_client}\\
		\mathrm{PS:   }\hspace{10pt}& \boldsymbol{\theta} = \frac{ \sum_{k\in \mathcal{K}}D_k \boldsymbol{\theta}_k  }{D}, \label{FL_PS}
		\end{align}
	\end{subequations}
}
	in which, each client optimizes its model parameters $\boldsymbol{\theta}_k$ based on $\mathcal{F}_k(\boldsymbol{\theta})$ as in (\ref{FL_client}) and transmits $\boldsymbol{\theta}_k$ to the PS, where the the model parameters are aggregated as in (\ref{FL_PS}). For an effective solution of   (\ref{FL_client}), gradient descent (GD) algorithm is used  iteratively such that an optimal local solution is obtained for each iteration. The PS and the clients exchange the updated model parameters until convergence~\cite{fl_spm_federatedLearning,robustFL,fl_By_Google}. In particular, the parameter update at the $t$th iteration is performed as 
	\begin{align}
	\label{eq:update1}
	\boldsymbol{\theta}_k^{(t+1)} = \boldsymbol{\theta}_k^{(t)} - \eta \mathbf{g}(\boldsymbol{\theta}_k^{(t)}),
	\end{align}
	where $\boldsymbol{\theta}_k^{(t)}$ denotes the model parameters at the $k$th client for the $t$th communication round/iteration with $t = 1,\dots,T$, $T$ is the total number of iterations, $\eta$ is the learning rate, and $\mathbf{g}(\boldsymbol{\theta}_k^{(t)})= \nabla \mathcal{F}_k(\boldsymbol{\theta}_k^{(t)})$ denotes the $P\times 1$ gradient vector, computed at the $k$th client, based on $\mathcal{D}_k$ and $\boldsymbol{\theta}_k^{(t)}$. Consequently, we can rewrite (\ref{eq:updatePS_Client_1}) as 
	\begin{subequations}
		\label{eq:updatePS_Client_1_Iterative}
		\begin{align}
		\mathrm{Client:   }\hspace{10pt}& \boldsymbol{\theta}_k^{(t+1)} = \boldsymbol{\theta}_k^{(t)} - \eta \mathbf{g}(\boldsymbol{\theta}_k^{(t)}), \hspace{10pt} k \in \mathcal{K}, \label{eq:updatePS_Client_1_Iterativeb}\\
		\mathrm{PS:   }\hspace{10pt}& \boldsymbol{\theta}^{(t+1)} = \frac{ \sum_{k\in \mathcal{K}}D_k \boldsymbol{\theta}_k^{(t+1)}  }{D}, \label{eq:updatePS_Client_1_Iterativea}
		\end{align}
	\end{subequations}
	which performs GD and iteratively reaches  convergence.

	\section{Hybrid Federated and Centralized Learning}
	\label{sec:HCFL}
	In this section, we introduce the proposed HFCL framework by taking into account the computational capability of the clients so that all clients can contribute to the learning task with their datasets regardless of their ability to compute the model parameters. The motivation is based on the following inadequacy observed in practice. In ML tasks, training a model requires huge computational power to compute the model parameters. This requirement cannot always be satisfied by the computational capabilities of the client devices. For this reason, an effective training of the ML model, we propose  a hybrid training framework that accounts for the computational capabilities of the clients. Specifically, we assume that only a portion of the clients with sufficient computational power performs FL, while the remaining clients, which suffer from inadequate computational capability, send their datasets to the PS for model computation, as illustrated in Fig.~\ref{fig_Diagc}.
	
	Let us define the set of clients who perform CL, i.e., sending datasets to the PS, and the ones who perform FL, i.e., transmitting the model parameters to the PS as $\mathcal{L} = \{1,\dots, L\}$ and $\bar{\mathcal{L}} = \{L +1,\dots, K\}$, respectively, where $\mathcal{K} = \mathcal{L} \bigcup \bar{\mathcal{L}}$ and $\mathcal{L} \cap \bar{\mathcal{L}} = \emptyset$. Furthermore, we refer to the clients in $\mathcal{L}$ and $\bar{\mathcal{L}}$ as \emph{inactive} and \emph{active} clients, respectively. {\color{black}The determination of client's condition (i.e., active/inactive) can be done by the devices sending information on their computational resources to the PS~\cite{FL_vehicular1,fl_clientSelection}.} {\color{black}By exploiting the availability of computational resources of the clients, the HFCL problem can be formulated as optimization of model parameters at the client and PS as follows 
	\begin{subequations}
		\label{eq:updatePS_Client_2}
		\begin{align}
		\mathrm{Client:   }\hspace{10pt}& \boldsymbol{\theta}_k = \arg \min_{\boldsymbol{\theta}} \mathcal{F}_k(\boldsymbol{\theta}), \hspace{10pt} k \in \bar{\mathcal{L}}, \label{FL_client2}\\
		\mathrm{PS:   }\hspace{10pt}& \boldsymbol{\theta}_k = \arg \min_{\boldsymbol{\theta}} \mathcal{F}_k(\boldsymbol{\theta}), \hspace{10pt} k \in \mathcal{L}, \label{FL_PS21} \\
		\mathrm{PS:   }\hspace{10pt}& \boldsymbol{\theta} = \frac{ \sum_{k\in \mathcal{K}}D_k \boldsymbol{\theta}_k  }{D}, \label{FL_PS22}
		\end{align}
	\end{subequations}}
	where the PS actively participates in the model computation process and computes $\boldsymbol{\theta}_k$ for $k \in \mathcal{L}$ since the inactive clients do not have sufficient computational capabilities. Once the model parameters are computed, the PS aggregates them as in (\ref{FL_PS22}). 
	
	\subsection{Noisy Learning Model}
	In HFCL, both the model parameters and the datasets are corrupted by noise due to wireless links~\cite{robustFL,elbir2020_FL_CE}. Compared to the noise in the datasets (for image classification, see Section~\ref{sec:sim}), the effect of noisy model parameters is significant since it directly corrupts the learning model~\cite{elbir2021HFCL}. During model transmission for $k \in \bar{\mathcal{L}}$, the noisy model parameters at the PS and the clients are respectively given by
	\begin{subequations}
		\label{eq:noisyModel}
		\begin{align}
		\mathrm{Client:   }\hspace{10pt}& \tilde{\boldsymbol{\theta}}_k  = \tilde{\boldsymbol{\theta}} + \Delta\tilde{\boldsymbol{\theta}}_k, k \in \bar{\mathcal{L}}, \label{noisyModelb}\\
		\mathrm{PS:   }\hspace{10pt}& \tilde{\boldsymbol{\theta}}  = \frac{1}{D} \bigg\{ \sum_{k \in \mathcal{L}} D_k \boldsymbol{\theta}_k  + \sum_{k \in \bar{\mathcal{L}}} D_k (\boldsymbol{\theta}_k + \Delta\boldsymbol{\theta}_k) \bigg\} . \label{noisyModela}
		\end{align}
	\end{subequations}
	In (\ref{noisyModela}), $\boldsymbol{\theta}_{k\in\bar{\mathcal{L}}}$ denotes the true model parameter transmitted from the active clients, while  $\Delta\boldsymbol{\theta}_{k\in \bar{\mathcal{L}}}$ represents the noise term added onto $\boldsymbol{\theta}_{k\in \bar{\mathcal{L}}}$, which is observed at the PS. After the  model aggregation in (\ref{noisyModela}), $\tilde{\boldsymbol{\theta}}$ is broadcast to the active clients and it is corrupted by the noise term  $\Delta\tilde{\boldsymbol{\theta}}_k$ as in (\ref{noisyModelb}) for $k \in \bar{\mathcal{L}}$. Notice that the first term on the right hand side of (\ref{noisyModela}) includes  no noise corruption since the model parameters are computed at the PS. Let us rewrite (\ref{noisyModela}) as 
	\begin{align}
	\label{eq5}
	\tilde{\boldsymbol{\theta}} = \frac{1}{D} \sum_{k \in \mathcal{K}} D_k \boldsymbol{\theta}_k +\frac{1}{D} \sum_{k\in \bar{\mathcal{L}}} D_k  \Delta\boldsymbol{\theta}_k.
	\end{align}
	Therefore, using (\ref{FL_PS}), (\ref{eq5}) becomes
	\begin{align}
	\label{noiseAtClient}
	\tilde{\boldsymbol{\theta}} = \boldsymbol{\theta} +  \widetilde{\Delta \boldsymbol{\theta}},
	\end{align}
	where $\widetilde{\Delta\boldsymbol{\theta}} = \frac{1}{D}\sum_{k \in \bar{\mathcal{L}}} D_k  \Delta\boldsymbol{\theta}_k$  is the aggregated noise term due to the model transmission from the active clients to the  PS. Without loss of generality, we assume that both noise terms due to model transmission in client-PS and PS-client links, i.e., $\widetilde{\Delta \boldsymbol{\theta}}$ and $\Delta \tilde{\boldsymbol{\theta}}_k$ $(k\in \bar{\mathcal{L}})$, are additive white Gaussian noise (AWGN) vectors~\cite{robustFL,elbir2020_FL_CE}{\color{black}\footnote{\color{black}While the Gaussian assumption is widely used in the relevant literature~\cite{elbir2020_FL_CE,convergenceFLAll,fl_convergenceOnNIIDData,robustFL}, it is worth noting that this assumption is not completely satisfied in practice due to the source/channel coding and quantization operations during the transmission of the model parameters. In addition, recent works, e.g.,~\cite{fl_noisyChanelsGunduz}, show that the noise on the model updates can also be modeled as unimodal symmetric distribution, which is close to Gaussian distribution.}}. Hence, regarding the aggregation noise at the PS, we have  $\mathbb{E}\{ \widetilde{\Delta \boldsymbol{\theta}}\} =\mathbf{0}$ and $\mathbb{E}\{\widetilde{\Delta \boldsymbol{\theta}} \widetilde{\Delta \boldsymbol{\theta}}^\textsf{T}\} = \tilde{\sigma}^2\mathbf{I}_P$.  Likewise, at the $k$th active client ($k\in \bar{\mathcal{L}}$), we have $\mathbb{E}\{ \Delta \tilde{\boldsymbol{\theta}}_k\}=\mathbf{0}$, $\mathbb{E}\{ \Delta \tilde{\boldsymbol{\theta}}_k \Delta\tilde{ \boldsymbol{\theta}}_k^\textsf{T}\} = {\sigma}_k^2\mathbf{I}_P$ and $\mathbb{E}\{ \Delta \boldsymbol{\theta}_{k_1} \Delta \boldsymbol{\theta}_{k_2}^\textsf{T}\} = \mathbf{0}$ for $k_1,k_2 \in \bar{\mathcal{L}}$ and $k_1 \neq k_2$. Furthermore, by using (\ref{noiseAtClient}) and  rewriting (\ref{noisyModelb}) as
	\begin{align}
	\label{noiseAtActiveClients}
	\tilde{\boldsymbol{\theta}}_k = \boldsymbol{\theta} + \widetilde{\Delta \boldsymbol{\theta}} + \Delta\tilde{\boldsymbol{\theta}}_k = \boldsymbol{\theta} + \widetilde{\Delta \boldsymbol{\theta}}_k,
	\end{align}
	we can define the noise on $\boldsymbol{\theta}$ at the $k$th client with variance $ \tilde{\sigma}^2 + \sigma_k^2$. Specifically, $\tilde{\sigma}^2$ corresponds to the noise term $\widetilde{\Delta\boldsymbol{\theta}}$ in (\ref{noiseAtClient}) generated during the model transmission from the active clients to the PS, while $\sigma_k^2$ corresponds to $\widetilde{\Delta{\boldsymbol{\theta}}}_k$ in (\ref{noiseAtActiveClients}), which is due to the transmission of the model parameters from the PS to the $k$th active client.

	Upon the above analysis on the noise corruption, it is clear that two different noise corruptions occur during training: noise corruption with $\tilde{\sigma}^2$ and $\tilde{\sigma}^2 + \tilde{\sigma}_k^2$ for inactive and active clients, respectively. In the sequel, we  modify the loss functions used in the training process according to the noise corruption to provide an expectation-based convergence~\cite{fl_By_Google,fl_convergenceOnNIIDData}.

	Let us first consider the loss function for active clients to solve (\ref{FL_client2}). In order to solve the local problem in (\ref{FL_client2}) effectively, the loss function should be modified to take into account the effect of noise.  Thus, we define a regularized loss function $\bar{\mathcal{F}}_k (\boldsymbol{\theta})$ as
	\begin{align}
	\label{lossFunctionModified}
	\bar{\mathcal{F}}_k(\boldsymbol{\theta}) = \mathcal{F}_k(\boldsymbol{\theta}) + (\tilde{\sigma}^2 + \sigma_k^2) ||\mathbf{g}(\boldsymbol{\theta}_k) ||^2,
	\end{align}
	where $\tilde{\sigma}^2 + \sigma_k^2$ describes the total noise term added onto $\boldsymbol{\theta}$ at the $k$th client ($k\in \bar{\mathcal{L}}$) in (\ref{noiseAtActiveClients}). The loss function in (\ref{lossFunctionModified}) is widely used in stochastic optimization~\cite{stochasticOpt} and it can be obtained via  a first-order Taylor  expansion of the expectation-based loss $\mathbb{E}\{|| \mathcal{F}_k(\boldsymbol{\theta} + \widetilde{\Delta \boldsymbol{\theta}}_k)||^2 \}$, which can be approximately written as 
	\begin{align}
	\mathbb{E}\{|| \mathcal{F}_k(\boldsymbol{\theta} + &\widetilde{\Delta \boldsymbol{\theta}}_k)||^2 \}\approx \mathbb{E}\{|| \mathcal{F}_k(\boldsymbol{\theta} ) + \widetilde{\Delta \boldsymbol{\theta}}_k\nabla\mathcal{F}_k(\boldsymbol{\theta} )     ||^2 \}, \nonumber \\
	&\approx \mathbb{E}\{|| \mathcal{F}_k(\boldsymbol{\theta} )||^2 \}  + \mathbb{E}\{||\widetilde{\Delta \boldsymbol{\theta}}_k||^2 \} \mathbb{E}\{|| \nabla\mathcal{F}_k(\boldsymbol{\theta} )     ||^2 \}, \nonumber \\
	& \approx \mathbb{E}\{|| \mathcal{F}_k(\boldsymbol{\theta} )||^2 \} +   (\tilde{\sigma}^2 + \sigma_k^2) ||\mathbf{g}(\boldsymbol{\theta}_k) ||^2, \label{costApprox}
	\end{align}
	where the first term in (\ref{costApprox}) corresponds to the minimization of the loss function with perfect estimation and the second term is the additional cost due to noise. The expectation-based loss in (\ref{costApprox}) provides a good approximation under the effect of uncertainties due to the noise by adding the regularizer term $(\tilde{\sigma}^2 + \sigma_k^2) ||\mathbf{g}(\boldsymbol{\theta}_k) ||^2$~\cite{stochasticOpt,robustFL}. Similarly, the loss function in (\ref{FL_PS21}) for inactive clients at the PS can be regularized as
	\begin{align}
	\label{lossPassive}
	\tilde{ \mathcal{F}}_k(\boldsymbol{\theta}) = \mathcal{F}_k(\boldsymbol{\theta}) + \tilde{\sigma}^2  ||\mathbf{g}(\boldsymbol{\theta}_k) ||^2.
	\end{align}
	The combination of  (\ref{noisyModela}), (\ref{lossFunctionModified}) and (\ref{lossPassive})  yields the regularized version of the HFCL problem as
	\begin{subequations}
		\label{eq:HFCL2}
		\begin{align}
		\mathrm{Client:   }\hspace{7pt}& \boldsymbol{\theta}_k = \arg \min_{\boldsymbol{\theta}}  \bar{ \mathcal{F}}_k(\boldsymbol{\theta}), \hspace{5pt} k \in \bar{\mathcal{L}}, \label{HFCL2c}\\
		\mathrm{PS:   }\hspace{7pt}& \boldsymbol{\theta}_k = \arg \min_{\boldsymbol{\theta}} \tilde{ \mathcal{F}}_k(\boldsymbol{\theta}), \hspace{10pt}k \in \mathcal{L}, \label{HFCL2a} \\
		\mathrm{PS:   }\hspace{7pt}& \tilde{\boldsymbol{\theta}}  = \frac{1}{D} \bigg\{ \sum_{k \in \mathcal{L}} D_k \boldsymbol{\theta}_k  + \sum_{k \in \bar{\mathcal{L}}} D_k (\boldsymbol{\theta}_k + \Delta\boldsymbol{\theta}_k) \bigg\} , \label{HFCL2b}
		\end{align}
	\end{subequations}
	where the training on the  active and inactive clients are performed in (\ref{HFCL2c}) and (\ref{HFCL2a}), respectively. To effectively solve (\ref{eq:HFCL2}), the GD algorithm is utilized to find an optimal\footnote{\color{black}The optimality of the proposed approach is subject to finding the minimizer of the loss function $\bar{\mathcal{F}}_k(\boldsymbol{\theta})$ at each iteration via the GD method as in the other FL works~\cite{fl_convergenceOnNIIDData,fl_noisyChanelsGunduz,fl_noisyChanels}. While the global optimality of GD can be guaranteed for shallow neural networks, e.g., with a single layer, it usually reaches to a local optimum for wide and deep learning models~\cite{deepLearningScience}.} model $\boldsymbol{\theta}$ for all clients. Then, the model parameter update is performed iteratively as follows
	\begin{subequations}
		\label{eq:HFCL3}
		\begin{align}
		\mathrm{Client:   }\hspace{10pt}& \boldsymbol{\theta}_k^{(t+1)} = \boldsymbol{\theta}_k^{(t)}- \eta  \bar{\mathbf{g}}_k(\boldsymbol{\theta}_k^{(t)}), \hspace{10pt} k \in \bar{\mathcal{L}}, \label{HFCL3c} \\
		\mathrm{PS:   }\hspace{10pt}& \boldsymbol{\theta}_k^{(t+1)} = \boldsymbol{\theta}_k^{(t)}- \eta  \tilde{\mathbf{g}}_k(\boldsymbol{\theta}_k^{(t)}), \hspace{10pt} k \in \mathcal{L}, \label{HFCL3a} \\
		\mathrm{PS:   }\hspace{10pt}& \tilde{\boldsymbol{\theta}}^{(t+1)}  = \frac{1}{D}  \sum_{k \in \mathcal{K}} D_k \boldsymbol{\theta}_k^{(t+1)} , \label{HFCL3b}
		\end{align}
	\end{subequations}
	where $ \bar{\mathbf{g}}_k(\boldsymbol{\theta}_k^{(t)}) = \nabla \bar{\mathcal{F}}_k(\boldsymbol{\theta}_k^{(t)})$ and  $ \tilde{\mathbf{g}}_k(\boldsymbol{\theta}_k^{(t)}) = \nabla \tilde{\mathcal{F}}_k(\boldsymbol{\theta}_k^{(t)})$ denote the gradient of the regularized loss function in (\ref{lossFunctionModified}) and (\ref{lossPassive}), respectively. In (\ref{eq:HFCL3}),  (\ref{HFCL3c}) and (\ref{HFCL3a}) compute the parameter updates for  active and inactive clients, respectively. Afterwards, the model aggregation is performed at the PS as in (\ref{HFCL3b}), and the aggregated model $\tilde{\boldsymbol{\theta}}^{(t+1)}$ is broadcast to the active clients.

	\subsection{Communication Delay During Model Training}
	The bandwidth resources need to be optimized to reduce the  latency of the transmission of  both $\boldsymbol{\theta}_k$ ($k\in \bar{\mathcal{L}}$) and $\mathcal{D}_k$ ($k\in \mathcal{L}$) to the PS during training. Let $\tau_k$ be the communication time for the $k$th client to transmit its either dataset ($\mathcal{D}_k$ for $k\in \mathcal{L}$) or model parameters ($\boldsymbol{\theta}_k$ for $k\in \bar{\mathcal{L}}$). We define 
	\begin{align}
	\label{latency}
	\tau_k = \frac{d_k}{R_k},
	\end{align}
	where $d_k$ denotes the number of dataset symbols to transmit and $R_k = B_k \ln (1 + \textsf{SNR}_k )$ is the achievable transmission rate. Herein,  $B_k$ and $\textsf{SNR}_k$ denote the allocated bandwidth and the signal-to-noise ratio (SNR) for the $k$th client, respectively. 	The PS solves $\min_{B_k} \max_{k\in \mathcal{K}} \hspace{6pt} \tau_k $ to optimize the bandwidth allocation by minimizing the maximum communication delay. This is because the model aggregation in the PS can be performed only after the completion of the slowest transmission for $k\in \mathcal{K}$.  Although $R_k$ can vary for $k\in \mathcal{K}$, $d_k$ differentiates more significantly than $R_k$ between the inactive (i.e., $k\in \mathcal{L}$) and active (i.e., $k\in \bar{\mathcal{L}}$) clients~\cite{latencyMinFL,HFL}. Especially, depending on the client type, $d_k$ can be given by
	\begin{align}
	\label{latency2}
	d_k = \left\{\begin{array}{ll}
	P, & k \in \bar{\mathcal{L}} \\
	\textsf{d}_k, & k\in \mathcal{L}
	\end{array}\right.,
	\end{align}
	which is fixed to the number of model parameters $P$ for the active clients, and to $\textsf{d}_k = D_k(U_xV_x + U_yV_y)$ for $D_k$ input ($\in \mathbb{R}^{U_x \times V_x}$) and output ($\in \mathbb{R}^{U_y \times V_y}$) dataset samples. 
	
	Since the dataset size is usually larger than the number of model parameters in ML applications, i.e., $\textsf{d}_{k\in \mathcal{L}} > P$~\cite{elbir2020FL_HB,FL_Gunduz,fl_spm_federatedLearning}, the dataset transmission of the inactive clients is expected to take longer than the model transmission of the active clients, i.e., $\tau_{k\in \mathcal{L}} > \tau_{k\in \bar{\mathcal{L}}}$~\cite{latencyMinFL}. Previous FL-based works reported that $\tau_{k\in \mathcal{L}}$ can be approximately $10$ times  longer than $\tau_{k\in \bar{\mathcal{L}}}$~\cite{elbir2020_FL_CE,elbir2020FL_HB}. This introduces a significant delay especially at the beginning of the training since the HFCL problem in (\ref{HFCL2a}) or (\ref{HFCL3a}) can be performed only if $\mathcal{D}_{k\in \mathcal{L}}$ is collected at the PS for the first iteration. 
	To tackle this issue and keep the training continuing, in the following, we propose two approaches, namely,  ICpC and SDT in Section~\ref{sec:HFCLicpc} and \ref{sec:HFCLsdt}, respectively. Both of these techniques are applied only at the beginning of the training to handle effectively the dataset transmission of the inactive clients.

	\begin{algorithm}[t]
		\begin{algorithmic}[1] 
			\caption{ \bf HFCL-ICpC}
			\Statex {\textbf{Input:} $\eta$, $\mathcal{D}_{k\in\mathcal{K}}$}. \label{alg:HFCL-ICpC}
			\Statex {\textbf{Output:} $\boldsymbol{\theta}$.}
			\State Initialize with $\boldsymbol{\theta}_{k}^{(t)}$ for $t=0$.
			\State \textbf{repeat} 
			\State \indent \textbf{if} $k \in \bar{\mathcal{L}}$ [Active clients],
			\State \indent\indent \textbf{if} $ t =0$
			\State\indent \indent\indent $t' := t$.
			\State \indent\indent\indent \textbf{repeat}
			\State \indent\indent\indent\indent $\boldsymbol{\theta}_k^{(t'+1)} = \boldsymbol{\theta}_k^{(t')}- \eta  \bar{\mathbf{g}}_k(\boldsymbol{\theta}_k^{(t')})$.
			\State \indent\indent\indent\indent $t' \leftarrow t' + 1$.
			\State \indent\indent\indent \textbf{until} $t' = N$
			\State\indent \indent\indent $\boldsymbol{\theta}_k^{(t+1)} = \boldsymbol{\theta}_k^{(N)}$.
			\State \indent\indent \textbf{else}
			\State \indent\indent\indent $\boldsymbol{\theta}_k^{(t+1)} = \boldsymbol{\theta}_k^{(t)}- \eta  \bar{\mathbf{g}}_k(\boldsymbol{\theta}_k^{(t)})$.
			\State \indent\indent \textbf{end}
			\State \indent\indent Send $\boldsymbol{\theta}_k^{(t+1)}$ to the PS.
			\State \indent\textbf{if} $k \in {\mathcal{L}}$ [Inactive clients],
			\State \indent\indent \textbf{if} $ t =0$
			\State\indent \indent \indent Send $\mathcal{D}_k $ to the PS.
			\State \indent\indent \textbf{else}
			\State \indent\indent\indent $\boldsymbol{\theta}_k^{(t+1)} = \boldsymbol{\theta}_k^{(t)}- \eta  \tilde{\mathbf{g}}_k(\boldsymbol{\theta}_k^{(t)})$.
			\State \indent\indent \textbf{end}
			\State \indent\textbf{end}
			\State \indent Aggregate models as ${\boldsymbol{\theta}}^{(t+1)}  = \frac{1}{D}  \sum_{k \in \mathcal{K}} D_k \boldsymbol{\theta}_k^{(t+1)}$.
			\State \indent Broadcast ${\boldsymbol{\theta}}^{(t+1)}$ to the clients for $k\in \bar{\mathcal{L}}$.
			\State \indent $t \leftarrow t + 1$.
			\State \textbf{until} convergence
		\end{algorithmic} 
	\end{algorithm}

	\section{HFCL With Increased Computation-per-Client}
	\label{sec:HFCLicpc}
	Prior to the model training, i.e., at $t=0$, the active clients need to wait until the inactive clients complete the dataset transmission, during which the active clients perform only one model computation. In order to keep the active clients continue model computing during the data transmission of inactive clients, we propose the ICpC approach, wherein the active clients move forward computing local model updates, but do  not send them to the PS until the dataset transmission of the inactive clients is completed~\cite{fl_By_Google}. This approach improves the convergence rate and the learning performance due to the continuation of the model updates at the inactive clients as similar observations are also reported in~\cite{fl_By_Google}, in which FL-only training is presented.

	The algorithmic steps of the HFCL-ICpC approach are presented in  Algorithm~\ref{alg:HFCL-ICpC}, for which the inputs are the datasets $\mathcal{D}_{k\in \mathcal{K}}$ and the learning rate $\eta$. Different from HFCL, HFCL-ICpC involves with the model updates at the active clients during first communication round, i.e., $t=0$, as in the lines $6-9$ of Algorithm~\ref{alg:HFCL-ICpC}. Using (\ref{latency}) and (\ref{latency2}), and defining $Q$ as the block size, the active clients can perform $N = \frac{\max_{k\in \mathcal{L}}\textsf{d}_{k}}{Q}$ iterations when $t=0$ until the dataset transmission of the inactive clients is completed (Line $10$ of Algorithm~\ref{alg:HFCL-ICpC}). Although this method keeps the active clients busy with model computation instead of staying idle, it does not reduce the communication latency at the first iteration. \textcolor{black}{As a result, the active clients perform $N$ local updates, which improve the convergence rate from $O(\frac{1}{t})$ to $O(\frac{N^2}{t})$ as compared to HFCL~\cite{convergenceFLAll}.} 

	\begin{algorithm}[t]
		\begin{algorithmic}[1] 
			\caption{\bf HFCL-SDT }
			\Statex {\textbf{Input:} $\eta$, $\mathcal{D}_{k \in \mathcal{K}}$, $\mathcal{F}_k$ as in (\ref{lossRegression})}. \label{alg:HFCL-SDT}
			\Statex {\textbf{Output:} $\boldsymbol{\theta}$}
			\State Initialize with $\boldsymbol{\theta}_{k}^{(t)}$ for $t=0$.
			\State \textbf{repeat} 
			\State \indent \textbf{if} $k \in \bar{\mathcal{L}}$ [Active clients],
			\State \indent \indent$\boldsymbol{\theta}_k^{(t+1)} = \boldsymbol{\theta}_k^{(t)}- \eta  \bar{\mathbf{g}}_k(\boldsymbol{\theta}_k^{(t)})$.
			\State \indent \indent Send $\boldsymbol{\theta}_k^{(t+1)}$ to the PS.
			\State \indent\textbf{if} $k \in {\mathcal{L}}$ [Inactive clients],
			\State \indent\indent \textbf{if} $ t \leq N$
			\State \indent \indent Send $\mathcal{D}_k^{(i)} $ to the PS for $i = (t-1)Q +1,\dots, tQ$.
			\State \indent \indent Use the collected dataset and compute  \par \indent\indent
			$\mathcal{F}_k(\boldsymbol{\theta}_k^{(t)}) = \large
			{  \frac{ \sum_{i=1}^{tQ} \| f( \mathcal{X}_k^{(i)}|\boldsymbol{\theta}_k^{(t)}) - \mathcal{Y}_k^{(i)}  \|^2 } { tQ} }$.
			\State \indent\indent \textbf{if} $ t > N$
			\State \indent \indent Use the  whole dataset and compute  \par \indent\indent $\mathcal{F}_k(\boldsymbol{\theta}_k^{(t)}) =\large
			{  \frac{ \sum_{i=1}^{D_k} \| f( \mathcal{X}_k^{(i)}|\boldsymbol{\theta}_k^{(t)}) - \mathcal{Y}_k^{(i)}  \|^2 } { D_k} }$.
			\State \indent\indent \textbf{end} 
			\State \indent \indent$ \boldsymbol{\theta}_k^{(t+1)} = \boldsymbol{\theta}_k^{(t)}- \eta  \tilde{\mathbf{g}}_k(\boldsymbol{\theta}_k^{(t)})$.
			\State \indent\textbf{end}
			\State \indent Aggregate models as ${\boldsymbol{\theta}}^{(t+1)}  = \frac{1}{D}  \sum_{k \in \mathcal{K}} D_k \boldsymbol{\theta}_k^{(t+1)}$.
			\State \indent Broadcast ${\boldsymbol{\theta}}^{(t+1)}$ to the clients for $k\in \bar{\mathcal{L}}$.
			\State \indent $t \leftarrow t + 1$.
			\State \textbf{until} convergence
		\end{algorithmic} 
	\end{algorithm}
	
	\section{HFCL With Sequential Data Transmission}
	\label{sec:HFCLsdt}
	Compared to HFCL and HFCL-ICpC, wherein the first communication round awaits for the completion of the transmission of inactive clients' datasets $\mathcal{D}_{k\in \mathcal{L}}$, HFCL-SDT involves with the transmission of smaller blocks of datasets. Hence,  the data transmission time per communication round is smaller while the total amount of time to transmit $\mathcal{D}_{k\in \mathcal{L}}$ to the PS is the same as in HFCL and HFCL-ICpC. 
	
	The algorithmic steps of the HFCL-SDT approach are given in Algorithm~\ref{alg:HFCL-SDT}. Let $Q$ be the block size of each portion. The number of blocks can be calculated as  $ N= \frac{\max_{k\in \mathcal{L}}\textsf{d}_k}{Q}$, which aims at minimizing the latency when $t=0$ by taking into account the largest dataset in terms of $\max_{k\in \mathcal{L}}\textsf{d}_k$ since we need to wait until the transmission of the largest dataset.  If $\textsf{d}_{k_1} = \textsf{d}_{k_2}$ for $k_1,k_2 \in \mathcal{L}$,  the size of the transmitted data for all clients becomes the same and equals to $Q$\footnote{As a special case, $Q=P$ can be selected to allow the transmitted data from both inactive and active clients is equal. However, this is only possible if $ N=\frac{\max_{k_\in \mathcal{L}} \textsf{d}_k}{P} < \gamma T$ so that dataset transmission is completed before $\gamma T$ iterations. If $\gamma =1$, then both dataset transmission and the training are completed at the same time, which is not efficient. Empirically, $\gamma=0.1$ is a good choice.}. If $N$ is not an integer, then $N=\lceil \frac{\max_{k\in \mathcal{L}}\textsf{d}_k}{Q} \rceil$ can be selected. Let us assume that $\mathcal{F}_k(\boldsymbol{\theta})$ is the regression loss as in (\ref{lossRegression})\footnote{The same expression can also be written for the loss function in (\ref{lossClassification}).}, then, the loss function for the inactive clients in the case of the HFCL-SDT algorithm are computed as
	\begin{align}
	\mathcal{F}_k(\boldsymbol{\theta}_k^{(t)})=\left\{ \hspace{-3pt}\begin{array}{ll}
	{  \frac{ \sum_{i=1}^{tQ} \| f( \mathcal{X}_k^{(i)}|\boldsymbol{\theta}_k^{(t)}) - \mathcal{Y}_k^{(i)}  \|^2 } { tQ} }, \hspace{5pt}t \leq N &   \\
	{  \frac{ \sum_{i=1}^{D_k} \| f( \mathcal{X}_k^{(i)}|\boldsymbol{\theta}_k^{(t)}) - \mathcal{Y}_k^{(i)}  \|^2 } { D_k} }, \hspace{5pt}t > N & \\ 
	\end{array}\right.\hspace{-13pt} ,
	\end{align}
	as given in lines $6-14$ of Algorithm~\ref{alg:HFCL-SDT}. The size of the training dataset, collected at the PS (i.e., $tQ$) for $\mathcal{F}_{k}(\boldsymbol{\theta}_k^{(t)})$, becomes larger as $t \rightarrow N$, and becomes equal to $D_k$ when $t=N$. When $t > N$, $\mathcal{F}_{k}(\boldsymbol{\theta}_k^{(t)})$ is computed for the whole dataset of inactive clients for $k\in \mathcal{L}$, as in the $11$th step of Algorithm~\ref{alg:HFCL-SDT}. While the size of the transmitted data from the inactive clients changes, the size of the transmitted information is fixed as $P$ for the active clients. This approach not only reduces the latency of the inactive clients, but also improves the learning accuracy as compared to HFCL since the features in the data are quickly learned at the PS due to the use of smaller datasets at the beginning. \textcolor{black}{As a result, the model parameters corresponding to the inactive clients are computed in a mini-batch learning way, in which the dataset is partitioned into mini-batches of size $Q$. Thus, the convergence rate of the inactive clients in HFCL-SDT is $O(\frac{1}{\sqrt{Qt}} + \frac{1}{t})$ while the rate of active clients is  $O(\frac{1}{t})$~\cite{miniBatchLearning}. }

	\section{Complexity Analysis and Communication Overhead}
	\label{sec:ConvergenceOverhead}
	In this section, we investigate the efficiency of the proposed HFCL framework in terms of complexity and communication overhead.
	{\color{black}
	\subsection{Complexity Analysis}
	The complexity of the learning schemes, i.e., CL, FL and HFCL, can be analyzed in terms of convergence speed since the computation of the model parameters is done by solving the optimization problems (\ref{lossML}), (\ref{eq:updatePS_Client_1}) and (\ref{eq:HFCL3}) via GD during training~\cite{convergenceGeoffrey,fl_convergenceOnNIIDData,robustFL}. Specifically, solving the optimization problems (\ref{lossML}), (\ref{eq:updatePS_Client_1}) and (\ref{eq:HFCL3}) for in CL, FL and HFCL, respectively, yields the same convergence rate of $O(1/t)$ although the accuracy of these algorithms differ depending on the noise corruption ($\tilde{\sigma}^2$ and $ \sigma_k^2$) on the model parameters~\cite{fl_convergenceOnNIIDData}. In particular, CL has higher learning accuracy than FL and HFCL since the model parameters are noise-free in CL while the accuracy of HFCL is between CL and FL since only a part of the model parameters (of inactive clients) are noise-free (see, e.g., Fig.~\ref{fig_acc_L}). Furthermore, the convergence rates of the proposed HFCL techniques, HFCL-ICpC and HFCL-SDT, are different from those of CL, FL and HFCL since they involve the participation of both active and inactive clients during training. In HFCL-ICpC, the convergence rate of solving the optimization problem for inactive clients is $O(\frac{1}{t})$ while it is  $O(\frac{N^2}{t})$ for the active clients due to performing $N$ local updates per iteration~\cite{fl_By_Google}. As a result, the convergence is faster in active clients as compared to the inactive ones. Also in HFCL-SDT, the convergence rates of the active and inactive clients are $O(\frac{1}{t})$ and $O(\frac{1}{\sqrt{Qt}}+\frac{1}{t})$, respectively, due to the dataset is partitioned into $Q$ mini-batches in inactive clients~\cite{miniBatchLearning}. Hence, one can conclude that the complexity of HFCL-ICpC is less than HFCL-SDT since the former reaches to convergence faster (see, e.g, Fig.~\ref{fig_acc_training}). We summarize the convergence rates of the HFCL methods for both active and inactive clients in Table~\ref{tableComparisonRate}.
	
 }

	{\color{black}Next, we investigate the convergence of the proposed HFCL framework. In the literature, the convergence of the ML models has been studied for centralized~\cite{optimizationForML} and federated~\cite{fl_convergenceOnNIIDData,robustFL} schemes separately. In this paper, we analyze the convergence of the hybrid scenario, in which the convergence of the loss functions for active $(\bar{\mathcal{F}}_k(\boldsymbol{\theta}))$ and inactive ($\tilde{\mathcal{F}}_k(\boldsymbol{\theta})$) clients in the presence of corrupted model parameters.} First, similar to the previous studies~\cite{fl_convergenceOnNIIDData,robustFL,fl_By_Google}, we make the following assumptions needed to ensure the convergence, which are typical for the $\l_2$-norm regularized linear regression, logistic regression, and softmax classifiers.
	
	\textit{Assumption 1:} The loss function $\mathcal{F}_k(\boldsymbol{\theta})$ is convex, i.e., $\mathcal{F}_k((1 - \lambda)\boldsymbol{\theta} + \lambda \boldsymbol{\theta}') \leq (1 - \lambda)\boldsymbol{\theta} \mathcal{F}_k(\boldsymbol{\theta}) + \lambda\mathcal{F}_k(\boldsymbol{\theta}')$ for $\lambda \in [0,1]$ and arbitrary  $\boldsymbol{\theta}$ and $\boldsymbol{\theta}'$.
	
	\textit{Assumption 2:} $\mathcal{F}_k(\boldsymbol{\theta})$ is \textit{L-Lipschitz}, i.e., $||\mathcal{F}_k(\boldsymbol{\theta}) - \mathcal{F}_k(\boldsymbol{\theta}')|| \leq  L ||\boldsymbol{\theta} - \boldsymbol{\theta}' ||$ for arbitrary  $\boldsymbol{\theta}$ and $\boldsymbol{\theta}'$.
	
	\textit{Assumption 3:} $\mathcal{F}_k(\boldsymbol{\theta})$ is \textit{$\beta$-Smooth}, i.e., $||\nabla\mathcal{F}_k(\boldsymbol{\theta}) - \nabla\mathcal{F}_k(\boldsymbol{\theta}')|| \leq  \beta ||\boldsymbol{\theta} - \boldsymbol{\theta}' ||$ for arbitrary  $\boldsymbol{\theta}$ and $\boldsymbol{\theta}'$.
	
	In order to prove the convergence of the loss functions for inactive and active clients, i.e., $\tilde{\mathcal{F}}_k(\boldsymbol{\theta})$ and $\bar{\mathcal{F}}_k(\boldsymbol{\theta})$, we first investigate the $\beta$-\textit{Smoothness} of $\bar{\mathcal{F}}_k(\boldsymbol{\theta})$ in  the following lemma.

	\begin{table}[t]
		
		\caption{\color{black}Convergence Rates for HFCL, HFCL-ICpC, and HFCL-SDT}
		\label{tableComparisonRate}
		\centering
		\begin{tabular}{c|c|c|c|c|c }
			\hline
			\hline
			\multicolumn{2}{c|}{HFCL} & \multicolumn{2}{|c|}{HFCL-ICpC} & \multicolumn{2}{|c}{HFCL-SDT} \\
			\hline
			Inactive & Active & Inactive & Active &Inactive & Active \\
			\hline 
			$O(\frac{1}{t})$ & $O(\frac{1}{t})$ &
			$O(\frac{1}{t})$ & $O(\frac{N^2}{t})$  &
			$O(\frac{1}{\sqrt{Qt}}\hspace{-3pt} + \hspace{-3pt} \frac{1}{t})$ & $O( \frac{1}{t})$\\
			\hline
			\hline
		\end{tabular}
	\end{table}
	
	\begin{lem} $\bar{\mathcal{F}}_k(\boldsymbol{\theta})$ is a $\bar{\beta}$-\textit{Smooth} function with $||\nabla \bar{\mathcal{F}}_k(\boldsymbol{\theta}) - \nabla\bar{\mathcal{F}}_k(\boldsymbol{\theta}')|| \leq  \bar{\beta} ||\boldsymbol{\theta} - \boldsymbol{\theta}' ||$, where $\bar{\beta} = (1 + \tilde{\sigma}^2 + \sigma_k^2) \beta$.
	\end{lem}
	\begin{IEEEproof} See Appendix~\ref{appendixLemma}.
	\end{IEEEproof}
	Using Lemma 1 and (\ref{lossPassive}), it is straightforward to show that $\tilde{\mathcal{F}}_k(\boldsymbol{\theta})$ is a  $\bar{\beta}$-\textit{smooth} function with  $\tilde{\beta} = \tilde{\sigma}^2\beta$.
	
	\begin{theo}
		At the $k$th active client ($k\in \bar{\mathcal{L}}$), the loss function $\bar{\mathcal{F}}_k(\boldsymbol{\theta})$  satisfies 
		\begin{align}
		\bar{\mathcal{F}}_k(\boldsymbol{\theta}^{(t)}) - \bar{\mathcal{F}}_k(\boldsymbol{\theta}^\star) \leq ||\boldsymbol{\theta}^{(0)} - \boldsymbol{\theta}^\star  ||^2  \frac{1}{2\eta  }  \frac{ 1}{t},
		\end{align}
		where the learning rate is subject to $\eta \leq \frac{1}{(1+\tilde{\sigma}^2 + \sigma_k^2) \beta }$, and $\boldsymbol{\theta}^\star$  is the minimizer of $\bar{\mathcal{F}}_k(\boldsymbol{\theta})$.
	\end{theo} 
	
	\begin{IEEEproof} See Appendix~\ref{appendix1}.
	\end{IEEEproof}

		Based on Theorem 1,  the loss function of the inactive clients $\tilde{\mathcal{F}}_k(\boldsymbol{\theta})$ is said to be convergent with  
	\begin{align}
	\tilde{\mathcal{F}}_k(\boldsymbol{\theta}^{(t)}) - \tilde{\mathcal{F}}_k(\boldsymbol{\theta}^\star) \leq ||\boldsymbol{\theta}^{(0)} - \boldsymbol{\theta}^\star  ||^2  \frac{1}{2\eta  }  \frac{ 1}{t},
	\end{align}
	where the learning rate obeys to $\eta \leq \frac{1}{(1+\tilde{\sigma}^2) \beta }$ since the problem for inactive clients is corrupted by the noise $\widetilde{\Delta\boldsymbol{\theta}} = \frac{1}{D}\sum_{k \in \bar{\mathcal{L}}} D_k  \Delta\boldsymbol{\theta}_k$ with variance $\tilde{\sigma}^2$ as in (\ref{noiseAtClient}).

	Consequently, if all  clients perform CL, i.e., $L=K$,  we have $\tilde{\sigma}^2 = \sigma_k^2 = 0$. Conversely, if all  clients perform FL, i.e., $L=0$,  the noise-free model parameters in (\ref{eq5}) will vanish and the noise variance becomes $\tilde{\sigma}_{\mathrm{FL}}^2= \frac{1}{D}\sum_{k \in \mathcal{K}} D_k\sigma_k^2 $. Thus, FL converges with lower learning performance than that of CL~\cite{fl_By_Google}. On the other hand, assuming $D_1 = \cdots=D_K$ and $\sigma_{1}^2 = \cdots = \sigma_{K}^2$,  the noise variance in HFCL $\tilde{\sigma}_{\mathrm{HFCL}}^2 = \frac{1}{D}\sum_{k \in \bar{\mathcal{L}}} D_k\sigma_k^2$ becomes less than $\tilde{\sigma}_{\mathrm{FL}}^2$ since $\sum_{k \in \bar{\mathcal{L}}} \sigma_k^2 < \sum_{k \in \mathcal{K}} \sigma_k^2$, where $K-L < K$ for $L\geq 1$. Therefore, the performance of HFCL is upper (lower) bounded by CL (FL) as $0\leq  \tilde{\sigma}_{\mathrm{HFCL}}^2 \leq \tilde{\sigma}_{\mathrm{FL}}^2$. 
	
	\textit{Remark 1:} If $D_{k_1} \neq D_{k_2}$, where $k_1,k_2\in \mathcal{K}$ and $k_1 \neq k_2$,  the performance of HFCL may be lower than FL. For instance, assuming that $\sigma_{1}^2 = \cdots = \sigma_{K}^2$, if $\frac{1}{D}\sum_{k \in \bar{\mathcal{L}}} D_k> \frac{1}{D}\sum_{k \in \mathcal{K}} D_k$,  HFCL converges with higher noise than that of FL since $ \tilde{\sigma}_{\mathrm{HFCL}}^2 > \tilde{\sigma}_{\mathrm{FL}}^2$. While the equality of $\sigma_{k\in \mathcal{K}}^2$ may be practical, the performance of HFCL and FL is subject to the size of the datasets~\cite{fl_By_Google,fl_convergenceOnNIIDData}.
	

	\subsection{Communication Overhead}
	\label{sec:Complexity}
	\textcolor{black}{The communication overhead is due to the transmission of the dataset or the model parameters that are exchanged between the clients and the PS, wherein several signal processing techniques, such as channel acquisition, quantization, and resource allocation, are employed. Hence,} the communication overhead can be measured by the number of transmitted symbols during model training~\cite{elbir2020FL_HB,FL_Gunduz,elbir2020cognitive,elbir2020_FL_CE}. Regarding the  communication overhead of CL ($\mathcal{T}_{\mathrm{CL}}$), it can be given by the number of symbols used to transmit datasets while the overhead during FL ($\mathcal{T}_{\mathrm{FL}}$) is proportional to the number of communication rounds $T$ and  model parameters $P$. Let ${\textsf{D}} = \sum_{k\in \mathcal{K}}{\textsf{d}}_k $ be the number of symbols of the whole dataset, then the communication overhead of CL, FL and HFCL are given respectively as
	\begin{align}
	\mathcal{T}_{\mathrm{CL}} &= {\textsf{D}}, \\
	\mathcal{T}_{\mathrm{FL}} &= 2TPK, \\
	\mathcal{T}_{\mathrm{HFCL}} &= L{\textsf{d}}_{k\in \mathcal{L}} + 2TP(K-L),
	\end{align}
	where  $\mathcal{T}_{\mathrm{HFCL}}$ includes the transmission of the dataset from $L$ inactive clients and the model parameters from  $K-L$ active clients. It is reasonable to assume that the size of the datasets is larger than the size of the model parameters, i.e., $\mathcal{T}_\mathrm{FL}\leq \mathcal{T}_\mathrm{CL}$~\cite{elbir2020FL_HB,FL_Gunduz,elbir2020cognitive,elbir2020_FL_CE,fl_spm_federatedLearning},  which gives $\mathcal{T}_{\mathrm{FL}}\leq \mathcal{T}_{\mathrm{HFCL}}\leq \mathcal{T}_{\mathrm{CL}}$. Notice that the communication overhead is equal for all HFCL techniques proposed in this work, since they all involve the same amount of dataset and model transmission.

	\begin{figure}[t]
		\centering
		{\includegraphics[draft=false,width=\columnwidth]{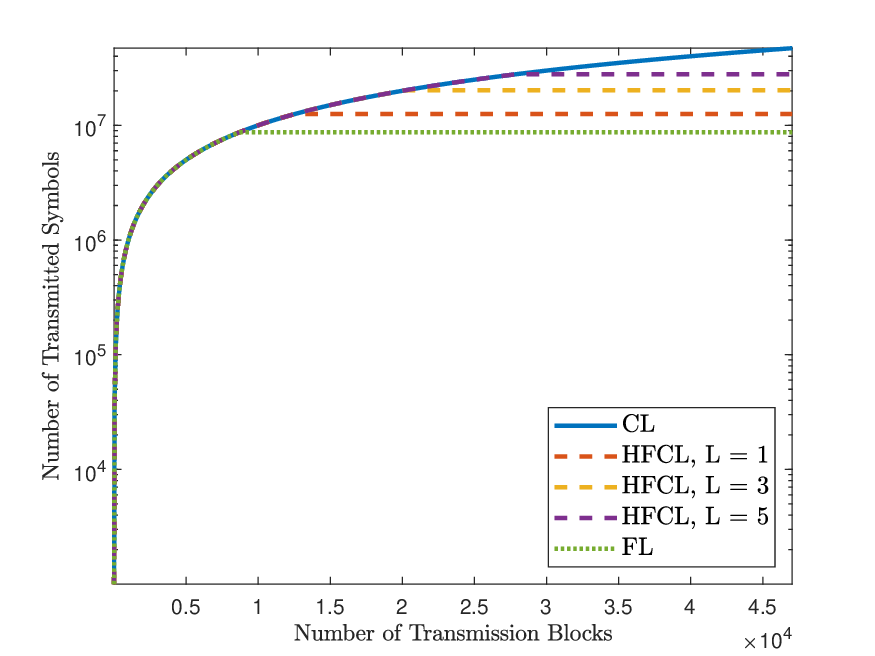} } 
		\caption{Communication overhead comparison.
		}
		\label{fig_TO}
	\end{figure}
	
	\begin{figure}[t]
		\centering
		{\includegraphics[draft=false,width=\columnwidth]{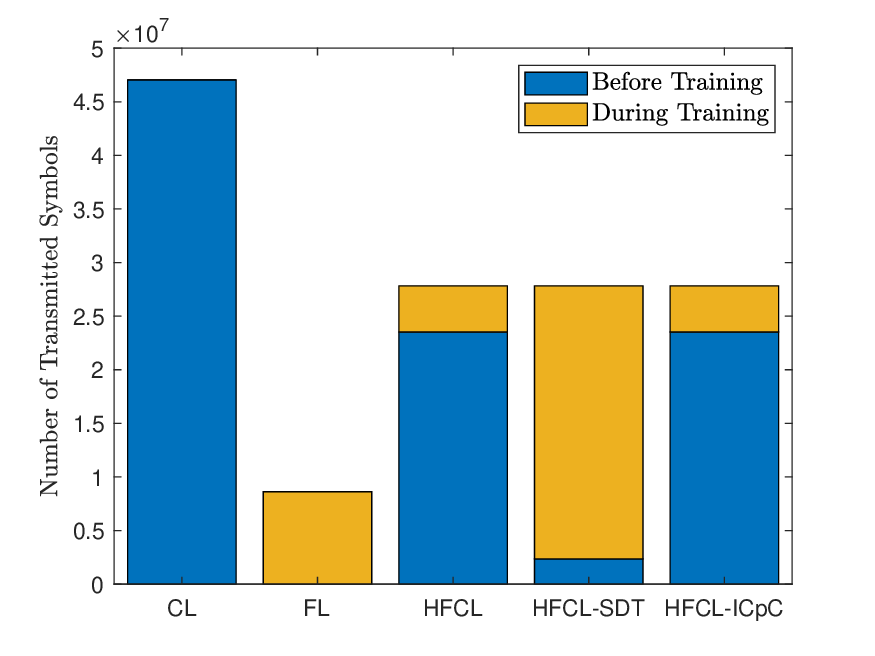} } 
		\caption{Number of transmitted symbols before and during training ($L=5$).
		}
		\label{fig_TO2}
	\end{figure}

	\section{Numerical Simulations}
	\label{sec:sim}
	In this section, we present the performance of the proposed HFCL frameworks in comparison to traditional CL- and FL-based training. We evaluate the performance on two datasets: i) image classification on the MNIST dataset~\cite{mnistlecun2010mnist} and ii) 3D object detection on the Lyft Level 5 AV dataset~\cite{lyft2019}.

	\subsection{Image Classification}
	We evaluate the performance on the MNIST dataset~\cite{mnistlecun2010mnist} including $28\times 28$ gray-scale images of handwritten digits with $10$ classes. The number of symbols on the whole dataset is ${\textsf{D}} = 28^2\cdot60,000 \approx 47\times 10^6$. During model training, the dataset is partitioned into $K=10$ blocks, each of which is available at the clients with identically independent distribution. Further, we train a CNN with $6$ layers. The first layer is $28\times 28$ input layer. The second and the fourth layers are convolutional layers with  $5\times 5$@$128$ and $3\times 3$@$128$ spatial filters, respectively. After each convolutional layer, there is a $\mathrm{ReLU}$ layer, which operates $\mathrm{max}(0, x)$ for its input $x$.  The output layer is a classification layer, which computes the probability distribution of the input data for $10$ classes. Thus, we have $P=128\cdot (5^2 + 3^2) = 4,352$ learnable parameters. The validation data of the MNIST dataset include $10,000$ images. The learning rate is selected as $0.001$, which is reduced by half after each $30$ iterations, and the mini-batch size is selected as $128$ for CL. The loss function was the cross-entropy cost as $ -\frac{1}{{{D}}} \sum_{i = 1}^{{{D}}} \sum_{c = 1}^{\bar{C}}  \bigg[ \mathcal{Y}_i^{(c)} \ln \hat{\mathcal{Y}}_i^{(c)} + (1 - \mathcal{Y}_i^{(c)}) \ln (1 - \hat{\mathcal{Y}}_i^{(c)})  \bigg],$	where $\{\mathcal{Y}_i^{(c)}, \hat{\mathcal{Y}}_i^{(c)}  \}_{i= 1, c=1}^{{{D}}, \bar{C}}$ are the true and predicted response for the classification layer  with $\bar{C}=10$. The classification accuracy is $	\mathrm{Accuracy}(\%) = \frac{\textsf{U}}{{D}}\times 100,$ in which the model identified the image class correctly $\textsf{U}$ times. Further, we define the SNR during model transmission as $\mathrm{SNR}_{\boldsymbol{\theta}} = 20\log_{10}\frac{||\boldsymbol{\theta}||_2^2}{ {\sigma}_{\boldsymbol{\theta}}^2  }  $, where ${\sigma}_{\boldsymbol{\theta}}^2 = \tilde{\sigma}^2 + \sigma_k^2$ denotes the total amount of noise variance during model transmission, for which we assume $\sigma_{L+1}^2 = \dots =\sigma_{K}^2$ for simplicity.

	\begin{figure}[t]
		\centering
		{\includegraphics[draft=false,width=\columnwidth]{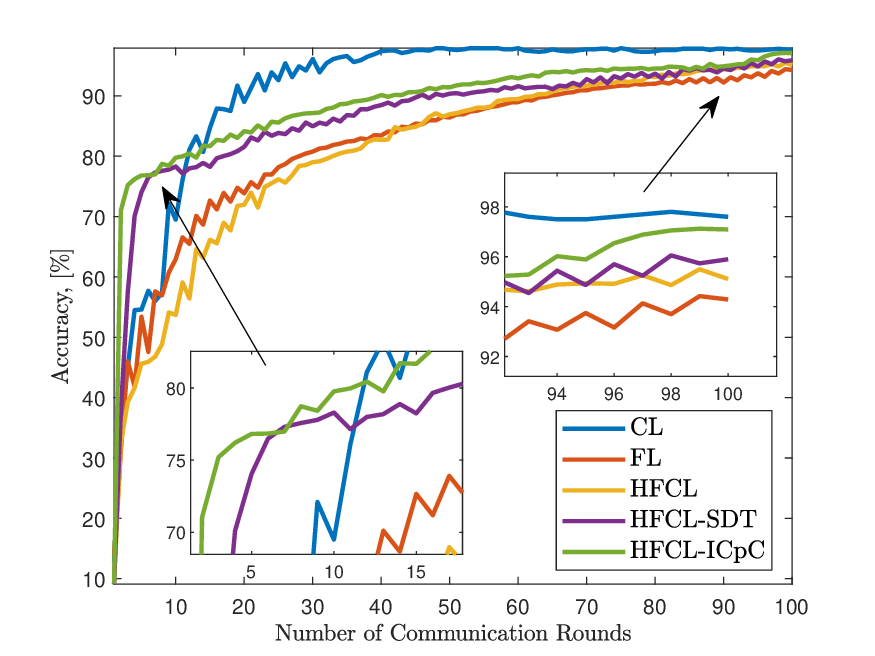} } 
		\caption{Classification accuracy versus the number of communication rounds  ($L=5$ $\mathrm{SNR}_{\boldsymbol{\theta}}=20$ dB, and  $B= 5$).
		}
		\label{fig_acc_training}
	\end{figure}
	
	\begin{figure}[t]
		\centering
		{\includegraphics[draft=false,width=\columnwidth]{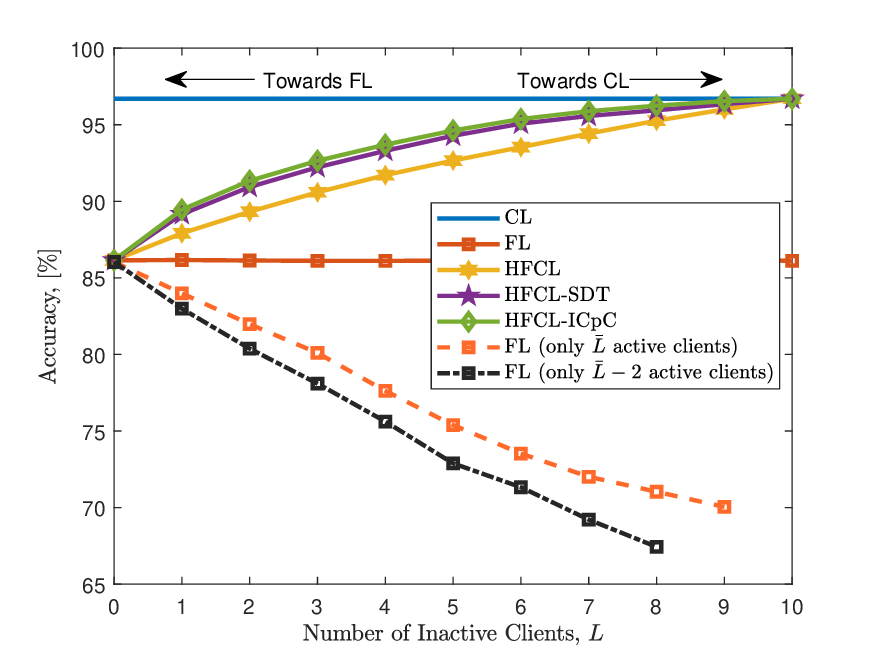} } 
		\caption{Classification accuracy versus  $L$  ($\mathrm{SNR}_{\boldsymbol{\theta}}=20$ dB, and  $B= 8$).
		}
		\label{fig_acc_L}
	\end{figure}

	Fig.~\ref{fig_TO} shows the communication overhead of CL, FL, and HFCL for  $L=\{0, 1,3,5,7,10\}$. During model training, we assume that $1000$ data symbols are transmitted at each transmission block. Thus, it takes approximately $47\times 10^3$ transmission blocks to complete CL-based training, while FL demands approximately  $8,5\times 10^3$ data blocks, which are approximately $6$ times lower  than that of CL. The communication overhead of HFCL (as well as HFCL-SDT and HFCL-ICpC) is between CL and FL since it depends on the number of inactive clients $L$ and approaches to $\mathcal{T}_\mathrm{CL}$ as $L\rightarrow K$.

	To explicitly present the overhead for the hybrid frameworks, the number of transmitted symbols before ($t=0$) and during ($t> 0$) training is presented in Fig.~\ref{fig_TO2}. Specifically, the whole communication overhead of CL is at $t=0$ before training due the dataset transmission. In contrast, the overhead of FL is observed during training since no dataset transmission is involved, yet model transmission is taken place when $t>0$. For the HFCL algorithms, they involve communication overheads before and during training since they involve both dataset and model transmission. Also, it is clear that the overhead of all hybrid frameworks are the same because they all involve the same amount of dataset and model transmission. However, they are distinguished in terms of the number of symbols transmitted before and during training. While HFCL and HFCL-ICpC have the same amount of overhead during training, HFCL-SDT has a lower overhead before training due to transmitting the datasets sequentially. Furthermore, we see that the overhead of FL is twice  the amount during training for HFCL and HFCL-ICpC. This is because $L=5$ for HFCL and HFCL-ICpC, and $L=K=10$ for FL. While the overhead of FL may seem lower compared to our HFCL frameworks, the former does not work if the clients have no capability for model computation, which is taken into account in the latter frameworks at the cost of a moderate increase in the overhead. 
	
	\begin{figure*}[t]
		\centering
		\subfloat[]{\includegraphics[draft=false,width=\columnwidth]{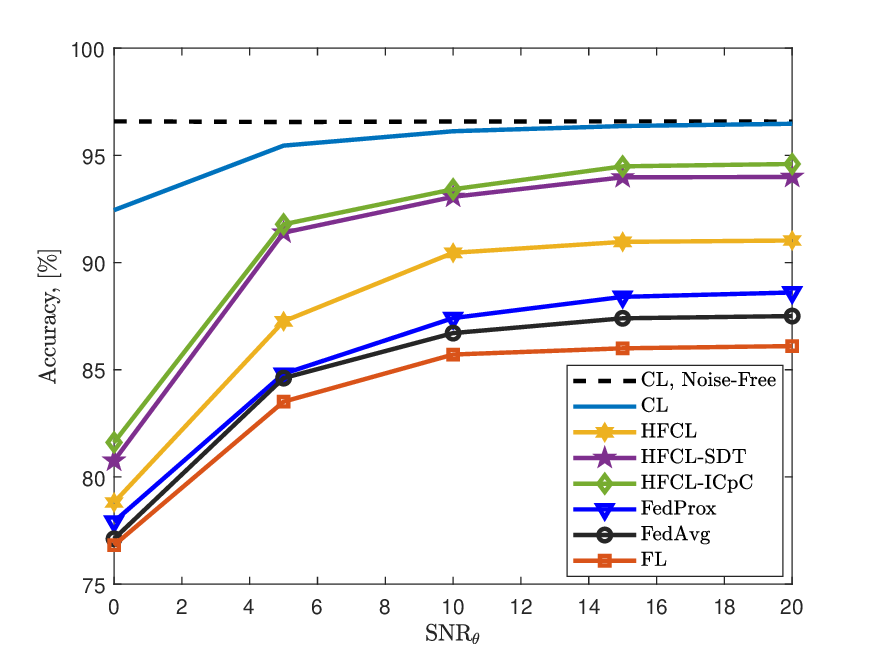} }  \subfloat[]{\includegraphics[draft=false,width=\columnwidth]{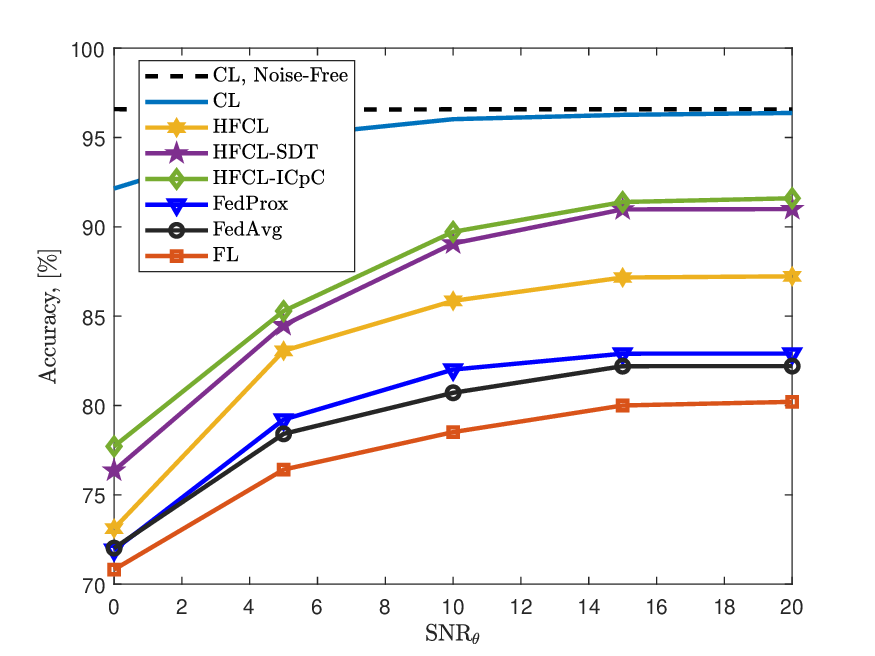} } 
		\caption{Classification accuracy versus $\mathrm{SNR}_{\boldsymbol{\theta}}$ for (a) i.i.d. and (b) non-i.i.d. datasets ($L=5$ and $B = 5$).
		}
		\label{fig_acc_SNRt}
	\end{figure*}

	\begin{figure}[t]
		\centering
		{\includegraphics[draft=false,width=\columnwidth]{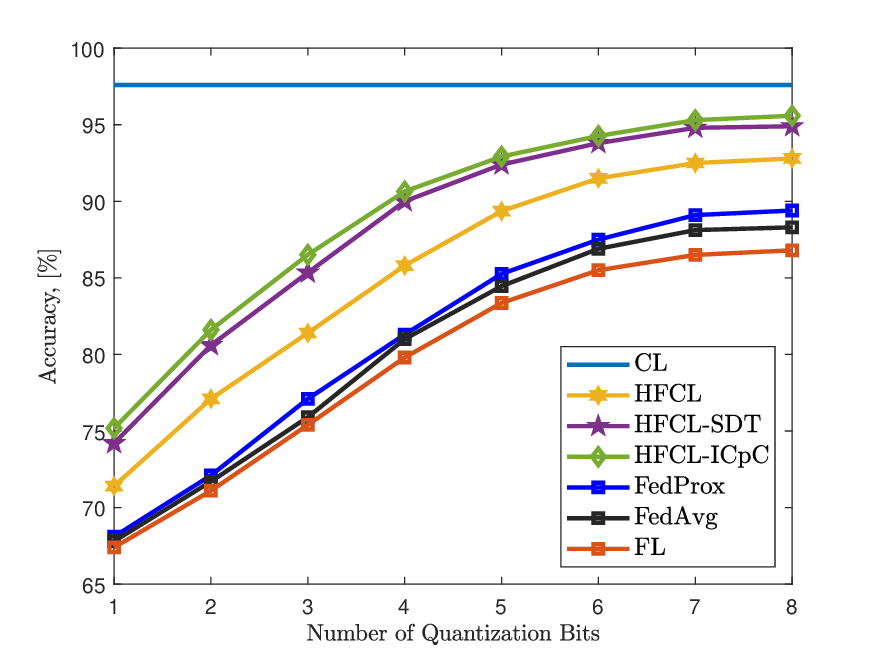} } 
		\caption{Classification accuracy versus $B$ ($\mathrm{SNR}_{\boldsymbol{\theta}} = 20$ dB).
		}
		\label{fig_acc_Q}
	\end{figure}

	\begin{figure*}[h]
		\centering
		\subfloat[]{\includegraphics[draft=false,width=.68\columnwidth]{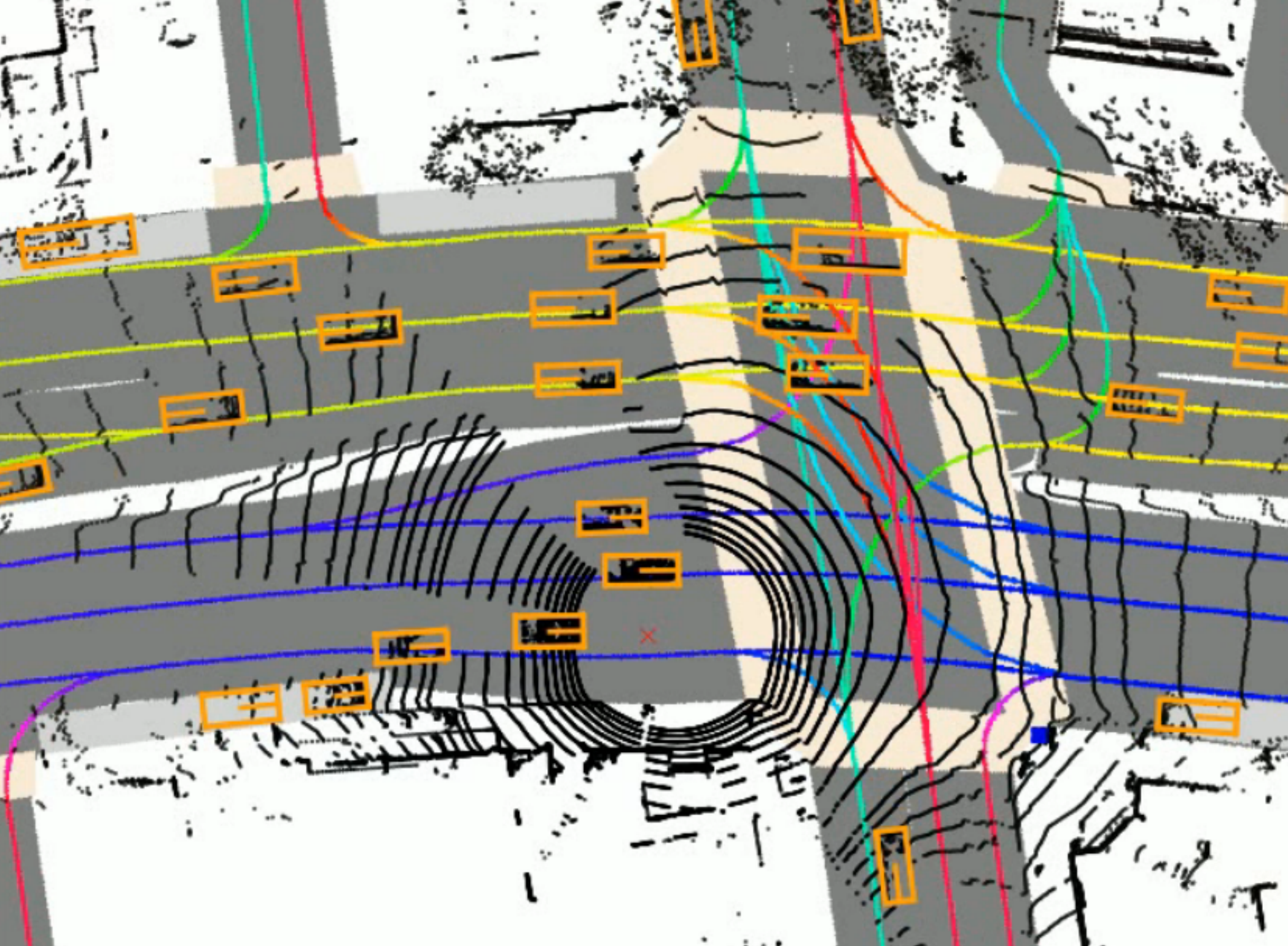} \label{fig_AVa}}
		\subfloat[]{\includegraphics[draft=false,width=.685\columnwidth]{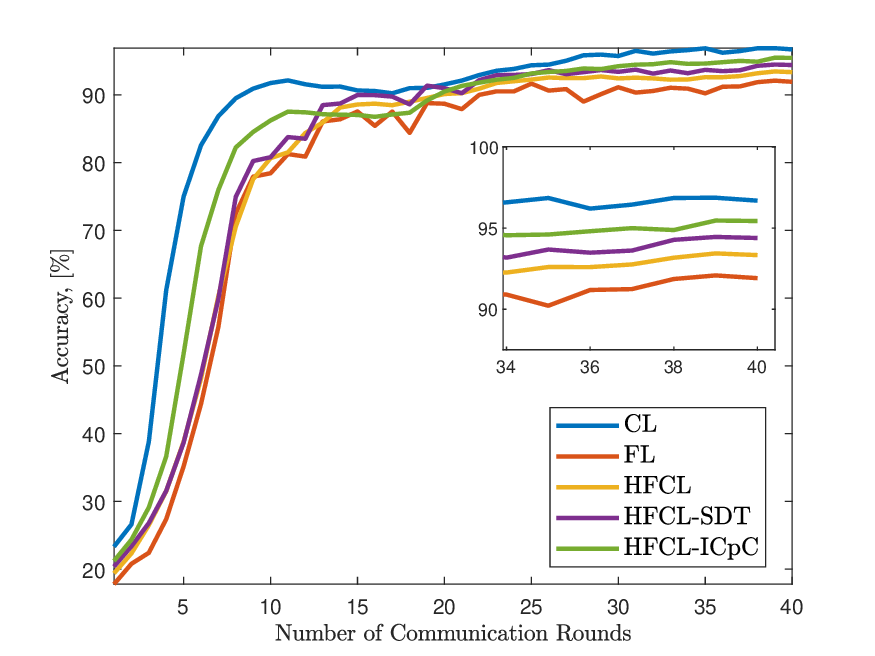} \label{fig_AVb}}
		\subfloat[]{\includegraphics[draft=false,width=.685\columnwidth]{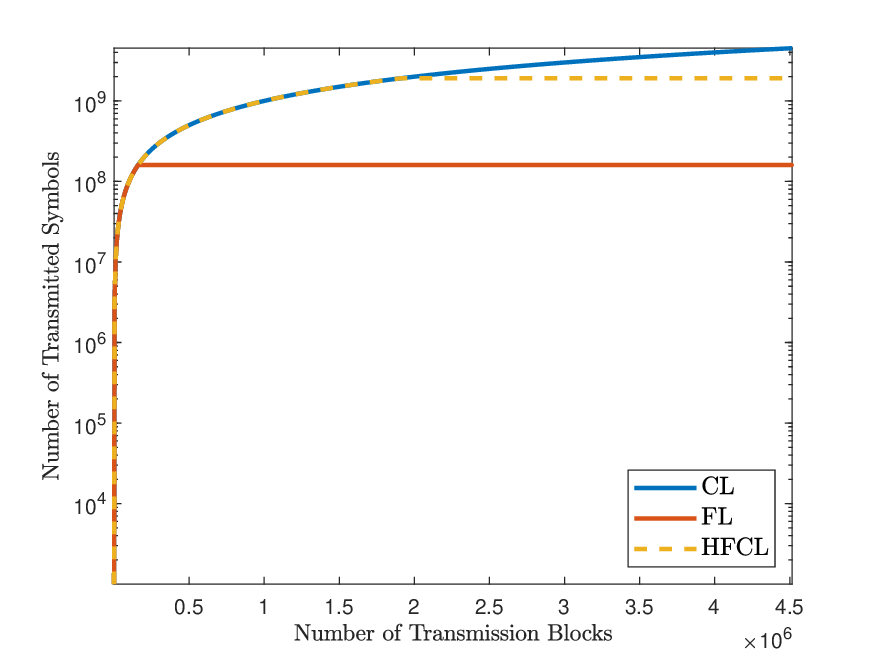} \label{fig_AVc}}
		\caption{3D object detection. (a) Visualization of input and output data, (b) object detection performance, and (c) communication overhead.
		}
		\label{fig_AV}
	\end{figure*}

	In Fig.~\ref{fig_acc_training}, we present the training performance of the competing methods for $L=5$, $\mathrm{SNR}_{\boldsymbol{\theta}}=20$ dB, and $B=5$ quantization bits {\color{black}with respect to number of communication rounds. Note that  we assume the dataset transmission is completed in CL prior to the training, and the completion of each communication round in FL corresponds to each iteration in CL.} HFCL-ICpC has a higher accuracy at the beginning of the  training due to multiple updates at the active clients, and it keeps outperforming HFCL and HFCL-SDT for the remaining iterations. Thanks to STD, HFCL-SDT also has higher performance than HFCL, which performs better than conventional FL. Notice that all the proposed HFCL approaches have moderate performance between FL and CL, as theoretically proved in Section~\ref{sec:ConvergenceOverhead}. We note that the performance of FL is computed by assuming that all $K$ clients participate in training, which is not possible if there exist inactive clients, for which FL is not the applicable training framework.

	Fig.~\ref{fig_acc_L} shows the classification accuracy with respect to the number of inactive clients $L$ when $\mathrm{SNR}_{\boldsymbol{\theta}}= 20$ dB. The proposed HFCL approaches perform better than FL for $0< L < K$ since, in FL, the collected models from the active clients are corrupted by the wireless channel whose effects reduce as $L \rightarrow K$.  When $L=0$, HFCL, HFCL-SDT, and HFCL-ICpC are identical to FL since all the clients are active whereas they perform the same as CL if $L=K=10$ since all the clients are inactive, i.e., they transmit their datasets to the PS, and the noise-free model parameters can be used for model training. Comparing the HFCL algorithms, it yields that	both HFCL-ICpC and HFCL-SDT provide higher accuracy than HFCL for $0< L < K$. HFCL-ICpC starts the training process when $t=0$ with locally-updated models at the active clients, thus, it provides a higher accuracy than both HFCL and HFCL-SDT. Also, HFCL-SDT starts the training with the same conditions as in HFCL while HFCL-SDT incorporates smaller datasets at the beginning for $t < N$. Thus,  higher accuracy levels are reached quicker than HFCL, which computes the model parameters  on the whole local dataset of the inactive clients for $t < N$, leading to a slower convergence rate.  
	
	{\color{black}In Fig.~\ref{fig_acc_L}, we also present the performance of FL with only $\bar{L}$ active clients where $\bar{L} = K-L$ is the number of active clients. That is to say, the learning model is trained on only the dataset of active clients, whereas it is tested on the whole dataset. We also consider the scenario that the training is conducted by only $\bar{L}-2$ active clients, two of which do not participate into the training due to the privacy concerns or limited conditions, such as having very low channel gain~\cite{fl_noisyChanelsGunduz,fl_clientScheduling}. We observe that FL becomes unable to learn the data as $L$ increases since the training is conducted only on the datasets of active clients. This shows the effectiveness of the proposed HFCL approach, in which all of the clients participate in the learning stage. It is worthwhile to note that when $L=K=10$, there will be no active clients, hence FL will not be performed. The performance loss can be severe  as $L\rightarrow K$ due to the absence of inactive clients' datasets  if the dataset is non-identically distributed because the active clients cannot learn the whole features in the dataset of other devices.}

	{\color{black}We further present the classification performance with respect to the noise corruption on the model parameters for both identically independent distribution  (IID) and non-IID cases in Fig.~\ref{fig_acc_SNRt}a and Fig.~\ref{fig_acc_SNRt}b, respectively. In IID scenario, the dataset is uniform randomly shuffled over the clients while for non-IID case, the dataset is sorted by labels and each client is then randomly assigned with 1 or 2 labels~\cite{fl_By_Google,fl_noisyChanels}. We compare the classification accuracy of the proposed HFCL methods with existing advanced FL techniques, i.e., {FedAvg}~\cite{fl_By_Google} and {FedProx}~\cite{fedprox}. Specifically, in FedAvg, all the clients  apply $N$ local model updates in each communication round whereas $N=1$ in conventional FL. Also, the clients employ different number of local model updates in FedProx (i.e., variable $N$)~\cite{fedprox}.} We set $\mathrm{SNR}_{\boldsymbol{\theta}}$ as $\mathrm{SNR}_{\boldsymbol{\theta}} = [0,20]$ dB and $\mathrm{SNR}_{\boldsymbol{\theta}} = \mathrm{SNR}_\mathcal{D}$, which is defined as the SNR of the noise added onto the transmitted datasets. \textcolor{black}{We can see from Fig.~\ref{fig_acc_SNRt} that the FL-based methods (e.g., FedAvg, FedProx and FL) are exposed to approximately $4.5\%$ loss in the learning accuracy while the HFCL methods perform $3.5\%$ poorer due to non-IID datasets. The slight performance improvement of HFCL methods can be attributed to the computation of model parameters at the PS. Furthermore, we observe that FedAvg has superior performance than FL due to multiple local model updates while FedProx performs better due to performing different number of local updates. Nevertheless, the proposed HFCL techniques have higher accuracy due to the computation of the model updates (of inactive clients) in the PS.} As expected, compared to the FL-based methods and the proposed hybrid algorithms, CL is more robust against noise because its model parameters are not corrupted by wireless transmission since they are computed at the PS. Comparing the effect of noise in datasets and model parameters, it can be concluded that the learning accuracy depends more on the corruptions in the model than the datasets. This is because the noise in the model makes it unable to learn the data while the noise in the dataset is, to some extend, helpful to make the model robust against the imperfections in the data. Therefore, artificial noise is usually added onto the dataset in many applications, such as image classification~\cite{deepLearningScience,survey_DL_Scalable}, and physical layer design in wireless communications~\cite{elbir2020_FL_CE,elbir2021FL_PHY}.
	
	{\color{black}In Fig.~\ref{fig_acc_Q}, we present the classification accuracy with respect to the quantization level of the model parameters for $B \in [1,8]$ when $\mathrm{SNR}_{\boldsymbol{\theta}}= 20$ dB. Note that the quantization is only applied to wireless-transmitted models, i.e., FL, FedAvg, FedProx and HFCL. Hence, the performance of CL does not change. The model parameters are quantized layer by layer between the maximum and minimum weights of each layer~\cite{elbirQuantizedCNN2019}.} As expected, the classification accuracy improves as $B$ increases due to the enhancement of the model precision. We can conclude that at least a $5$-bit quantization is required for a reliable classification accuracy.

	\subsection{3D Object Detection}
	We evaluate the performance of the HFCL framework on the 3D object detection problem in vehicular networks, based on the Lyft Level 5 AV dataset~\cite{lyft2019}, collected from lidar and cameras mounted on vehicles~\cite{elbir2020federated}. The input data are selected as a top view image of the ego vehicle, which includes the received lidar signal strengths for different elevations, as shown in Fig.~\ref{fig_AVa}. The output data are the classified representation of the vehicles/objects as boxes, which are obtained by the preprocessing of the images from the cameras, as illustrated in Fig.~\ref{fig_AVa}~\footnote{While we assumed that the dataset is labeled, the annotation of the objects in the data samples is an important task, for which the reader can refer to~\cite{elbir2020federated,elbir2020cognitive}.}. The training dataset is collected from $10$ vehicles in different areas after preprocessing of the camera and lidar data. We assume that  $L=3$ of the vehicles are inactive while the remaining ones are active clients. Each dataset includes $10^3$ input-output pairs, whose sizes are $336\times 336\times3$ and $336\times 336\times1$, respectively. Hence, the total number of data symbols are  $(336\times 336\times3 + 336\times 336)\times 10^4 \approx 4.5\times 10^9$. The dataset has $9$ classes, i.e., $\mathrm{car}$, $\mathrm{motorcycle}$, $\mathrm{bus}$, $\mathrm{bicycle}$, $\mathrm{truck}$, $\mathrm{pedestrian}$, $\mathrm{other}\hspace{2pt} \mathrm{vehicle}$, $\mathrm{animal}$, and $\mathrm{emergency}\hspace{2pt} \mathrm{vehicle}$, which are represented by the boxes as shown in  Fig.~\ref{fig_AVa}.  We have used the  U-net~\cite{unet} with $8$ convolutional layers to learn the features in the input data and achieve 3D object detection and segmentation.  The total number of parameters in U-net is approximately $2\times 10^6$, and the training is conducted for $T=40$ communication rounds.
	
	The training performance of the competing methods is presented in Fig.~\ref{fig_AVb}, from which we obtain similar observations as in the image classification scenario in the previous part. HFCL methods provide a moderate performance between CL and FL. Nevertheless, all of the vehicles can participate in training while the conventional FL methods cannot support it and the communication overhead of CL is prohibitive. Thus, HFCL provides a solution for the the trade-off between the computational capabilities of the clients and the communication overhead. 
	
	In Fig.\ref{fig_AVc}, the numbers of transmitted data symbols for CL, FL and HFCL are presented. The overhead of CL is due to the transmission of whole data symbols, i.e., approximately $ 4.5\times 10^9$. In contrast, the complexity of FL is due to the two way (edge $\leftrightarrows$ server) transmission of the model updates  during training until convergence, i.e., $2\times40\times  (2\times 10^6) = 160\times 10^6$ for $40$ iterations. As a result, FL and HFCL have approximately $28$ and $3$ times lower communication overhead compared to CL, respectively. The effectiveness of HFCL in terms of overhead is reduced compared to the image classification application because of the larger input size of the object detection problem (i.e., $336\times 336\times 3$ versus $28\times 28\times 1$).

	\section{Summary}
	\label{sec:conc}
	In this paper, we introduced a hybrid federated and centralized learning (HFCL) approach for distributed ML tasks. The proposed approach is helpful if a portion of the edge devices do not have the computational capability for model computation during  training, which is quite common in practice. In order to train the learning model collaboratively, only the active devices, which have sufficient computational capability, perform FL by computing the model updates on their local datasets whereas the remaining inactive devices, which do not have enough computational power, resort to CL and send their local datasets to the PS, which performs the model computation on behalf of them. As a result, HFCL provides a novel solution achieving a trade-off between FL and CL in terms of computation capability of edge devices and communication overhead. Moreover, the transmission of local datasets may generate delays during training if the dataset size of the inactive clients is large. We suggested a mitigation of this  problem by proposing HFCL-ICpC and HFCL-SDT frameworks. While the former improves the learning accuracy with a model computation-per-client, the latter reduces the size of the transmitted datasets. Compared to CL, the proposed approach is more advantageous due to the reduction in the communication latency. In the meantime,  compared to FL, a  drawback of the proposed HFCL approach is that HFCL requires the dataset transmission of the inactive clients, which may raise privacy concerns, however, these are already apparent in CL. Nevertheless, HFCL provides access to the dataset of all devices regardless of their computational capability whereas, in FL, only the dataset of the inactive clients could be used. We showed that our HFCL approach has a significant performance improvement as compared to FL-based training with only active clients since FL cannot access the dataset of all clients. As a future work, we plan to study the application of HFCL to physical layer applications, such as channel estimation, resource allocation, and beamforming. Furthermore, the noisy learning model may not represent the practical case, hence requires further research on realistic modeling.

	\appendices
	\section{Proof of Lemma 1}
	\label{appendixLemma}
	Using (\ref{lossFunctionModified}), we get
	\begin{align}
	&||\nabla \bar{\mathcal{F}}_k(\boldsymbol{\theta}) - \nabla\bar{\mathcal{F}}_k(\boldsymbol{\theta}')||  \nonumber \\
	&= || \nabla (\mathcal{F}_k(\boldsymbol{\theta})+ (\tilde{\sigma}^2 + \sigma_k^2)||\nabla \mathcal{F}_k(\boldsymbol{\theta}) ||^2 ) \nonumber \\
	& \hspace{50pt} - \nabla (\mathcal{F}_k(\boldsymbol{\theta}')+ (\tilde{\sigma}^2 + \sigma_k^2)||\nabla \mathcal{F}_k(\boldsymbol{\theta}')||^2)  ||\nonumber \\
	&=||  \big(\nabla \mathcal{F}_k(\boldsymbol{\theta}) + (\tilde{\sigma}^2 + \sigma_k^2)\nabla||\nabla \mathcal{F}_k(\boldsymbol{\theta}) ||^2 \big) \nonumber \\
	&\hspace{50pt}- \big(\nabla \mathcal{F}_k(\boldsymbol{\theta}') + (\tilde{\sigma}^2 + \sigma_k^2)\nabla||\nabla \mathcal{F}_k(\boldsymbol{\theta}') ||^2\big)  || \nonumber \\
	& =|| \nabla {\mathcal{F}}_k(\boldsymbol{\theta}) - \nabla{\mathcal{F}}_k(\boldsymbol{\theta}')  + (\tilde{\sigma}^2 + \sigma_k^2) \nonumber \\
	&\hspace{0pt} \times \big(\nabla \mathrm{tr}\{\nabla{\mathcal{F}}_k(\boldsymbol{\theta})^\textsf{T} \nabla {\mathcal{F}}_k(\boldsymbol{\theta})   \}  - \nabla \mathrm{tr}\{\nabla{\mathcal{F}}_k(\boldsymbol{\theta}')^\textsf{T} \nabla {\mathcal{F}}_k(\boldsymbol{\theta}')   \}\big)   || \nonumber \\
	& = ||  \nabla {\mathcal{F}}_k(\boldsymbol{\theta}) -\nabla {\mathcal{F}}_k(\boldsymbol{\theta}')   \nonumber \\
	&\hspace{50pt} + (\tilde{\sigma}^2 + \sigma_k^2) \big(   \nabla {\mathcal{F}}_k(\boldsymbol{\theta}) - \nabla{\mathcal{F}}_k(\boldsymbol{\theta}')  \big)     || \nonumber \\
	& = || (1 + \tilde{\sigma}^2 + \sigma_k^2) \big(\nabla {\mathcal{F}}_k(\boldsymbol{\theta}) -\nabla {\mathcal{F}}_k(\boldsymbol{\theta}')\big) || \nonumber \\
	&=(1 + \tilde{\sigma}^2 + \sigma_k^2) || \nabla {\mathcal{F}}_k(\boldsymbol{\theta}) -\nabla {\mathcal{F}}_k(\boldsymbol{\theta}') ||.\label{eq6}
	\end{align}
	By incorporating (\ref{eq6}), Assumption 2 and the fact that $1 + (\bar{\sigma}^2 + \sigma_k^2) \geq 0$, we get
	\begin{align}
	\label{lemmaResult}
	||\nabla \bar{\mathcal{F}}_k(\boldsymbol{\theta}) - \nabla\bar{\mathcal{F}}_k(\boldsymbol{\theta}')|| \leq  \tilde{\beta} ||\boldsymbol{\theta} - \boldsymbol{\theta}' ||^2,
	\end{align}
	where $\bar{\beta} = (1 + (\tilde{\sigma}^2 + \sigma_k^2)) \beta$.
	
	\section{Proof of Theorem 1}
	\label{appendix1}
	Using (\ref{lemmaResult}),  Assumption 2 and Assumption 3 imply that $\bar{\mathcal{F}}_k(\boldsymbol{\theta})$ is second order differentiable as $\nabla^2 \bar{\mathcal{F}}_k(\boldsymbol{\theta})\preceq \bar{\beta} \mathbf{I}_P $. Using this fact, performing a quadratic expression around $\bar{\mathcal{F}}_k(\boldsymbol{\theta})$ yields  
	\begin{align}
	\label{eq:quadraticExp}
	\bar{\mathcal{F}}_k(\boldsymbol{\theta}') &\leq \bar{\mathcal{F}}_k(\boldsymbol{\theta}) + \nabla \bar{\mathcal{F}}_k(\boldsymbol{\theta})^\textsf{T} (\boldsymbol{\theta}' - \boldsymbol{\theta})  + \frac{1}{2} \nabla^2 \bar{\mathcal{F}}_k(\boldsymbol{\theta}) ||\boldsymbol{\theta}' - \boldsymbol{\theta} ||^2 \nonumber \\
	&\leq \mathcal{F}_k(\boldsymbol{\theta}) + \nabla \bar{\mathcal{F}}_k(\boldsymbol{\theta})^\textsf{T} (\boldsymbol{\theta}' - \boldsymbol{\theta})  + \frac{1}{2} \bar{\beta}  ||\boldsymbol{\theta}' - \boldsymbol{\theta} ||^2.
	\end{align}
	Substituting the GD update $ \boldsymbol{\theta}' = \boldsymbol{\theta} - \eta \nabla \bar{\mathcal{F}}_k(\boldsymbol{\theta}) $ in (\ref{eq:quadraticExp}), we get
	\begin{align}
	\bar{\mathcal{F}}_k(\boldsymbol{\theta}') &\leq \bar{\mathcal{F}}_k(\boldsymbol{\theta}) + \nabla \bar{\mathcal{F}}_k(\boldsymbol{\theta})^\textsf{T} (\boldsymbol{\theta}' - \boldsymbol{\theta})  + \frac{1}{2} \bar{\beta}  ||\boldsymbol{\theta}' - \boldsymbol{\theta} ||^2 \nonumber \\
	& = \bar{\mathcal{F}}_k(\boldsymbol{\theta})  + \nabla \bar{\mathcal{F}}_k(\boldsymbol{\theta})^\textsf{T} (\boldsymbol{\theta} - \eta \nabla \bar{\mathcal{F}}_k(\boldsymbol{\theta})  - \boldsymbol{\theta}) \nonumber \\
	& \hspace{10pt}+ \frac{1}{2} \nabla^2 \bar{\mathcal{F}}_k(\boldsymbol{\theta}) ||\boldsymbol{\theta} - \eta \nabla \bar{\mathcal{F}}_k(\boldsymbol{\theta}) - \boldsymbol{\theta} ||^2 \nonumber \\
	& = \bar{\mathcal{F}}_k(\boldsymbol{\theta})  - \eta \nabla \bar{\mathcal{F}}_k(\boldsymbol{\theta})^\textsf{T} \nabla\bar{\mathcal{F}}_k(\boldsymbol{\theta}) +  \frac{1}{2} \bar{\beta}  ||\eta \nabla \bar{\mathcal{F}}_k(\boldsymbol{\theta}) ||^2  \nonumber \\
	&= \bar{\mathcal{F}}_k(\boldsymbol{\theta})  - \eta || \nabla \bar{\mathcal{F}}_k(\boldsymbol{\theta})||^2 + \frac{1}{2} \bar{\beta} \eta^2 || \nabla \bar{\mathcal{F}}_k(\boldsymbol{\theta})||^2 \nonumber \\
	& = \bar{\mathcal{F}}_k(\boldsymbol{\theta}) - (1 - \frac{\bar{\beta} \eta }{2}) \eta  || \nabla \bar{\mathcal{F}}_k(\boldsymbol{\theta})||^2, \label{useConvexity}
	\end{align}
	which bounds the GD update $\bar{\mathcal{F}}_k(\boldsymbol{\theta}')$ with $\bar{\mathcal{F}}_k(\boldsymbol{\theta})$. Now, let us bound $\bar{\mathcal{F}}_k(\boldsymbol{\theta}')$ with the optimal objective value $\bar{\mathcal{F}}_k(\boldsymbol{\theta}^\star)$. Using Assumption 1, we have
	\begin{align}
	\label{dueToConvexity}
	\bar{\mathcal{F}}_k(\boldsymbol{\theta}^\star )& \geq \bar{\mathcal{F}}_k(\boldsymbol{\theta})  + \nabla \bar{\mathcal{F}}_k(\boldsymbol{\theta})^\textsf{T} (\boldsymbol{\theta}^\star - \boldsymbol{\theta}), \nonumber \\
	\bar{\mathcal{F}}_k(\boldsymbol{\theta} )& \leq  \bar{\mathcal{F}}_k(\boldsymbol{\theta}^\star)  + \nabla \bar{\mathcal{F}}_k(\boldsymbol{\theta})^\textsf{T} (\boldsymbol{\theta} - \boldsymbol{\theta}^\star).
	\end{align}
	Furthermore, using $\eta \leq \frac{1}{\bar{\beta}}$, we have $-(1 - \frac{\bar{\beta}\eta}{2}) = \frac{1}{2}\bar{\beta}\eta - 1 \leq \frac{1}{2 }\bar{\beta} (1/\bar{\beta}) -1 = \frac{1}{2} - 1 = -\frac{1}{2}$. Thus, (\ref{useConvexity}) becomes
	\begin{align}
	\label{useConvexity2}
	\bar{\mathcal{F}}_k(\boldsymbol{\theta}') \leq \bar{\mathcal{F}}_k(\boldsymbol{\theta}) - \frac{\eta}{2}  || \nabla \bar{\mathcal{F}}_k(\boldsymbol{\theta})||^2
	\end{align}
	By plugging (\ref{dueToConvexity}) into (\ref{useConvexity2}), we get 
	\begin{align}
	&\bar{\mathcal{F}}_k(\boldsymbol{\theta}') \leq \bar{\mathcal{F}}_k(\boldsymbol{\theta}^\star) + \nabla \bar{\mathcal{F}}_k(\boldsymbol{\theta})^\textsf{T} (\boldsymbol{\theta} 
	- \boldsymbol{\theta}^\star ) - \frac{\eta}{2}  || \nabla \bar{\mathcal{F}}_k(\boldsymbol{\theta})||^2,
	\end{align}
	which can be rewritten as
	\begin{align}
	\label{eq1}
	\bar{\mathcal{F}}_k(\boldsymbol{\theta}') - &\bar{\mathcal{F}}_k(\boldsymbol{\theta}^\star) \leq 
	\nonumber\\
	&\frac{1}{2\eta} \bigg( 2\eta \nabla \bar{\mathcal{F}}_k(\boldsymbol{\theta})^\textsf{T} (\boldsymbol{\theta} 
	- \boldsymbol{\theta}^\star ) - \eta^2 || \nabla \bar{\mathcal{F}}_k(\boldsymbol{\theta})||^2  \bigg).
	\end{align}
	By adding $\frac{1}{2\eta}(||\boldsymbol{\theta} - \boldsymbol{\theta}^\star   ||^2 - ||\boldsymbol{\theta} - \boldsymbol{\theta}^\star   ||^2)$ into the right hand side of (\ref{eq1}), we get
	\begin{align}
	\label{eq2}
	\bar{\mathcal{F}}_k(\boldsymbol{\theta}') - &\bar{\mathcal{F}}_k(\boldsymbol{\theta}^\star) \leq 
	\nonumber\\
	&\frac{1}{2\eta} \bigg( ||\boldsymbol{\theta} - \boldsymbol{\theta}^\star   ||^2  - ||\boldsymbol{\theta} - \boldsymbol{\theta}^\star - \eta \nabla \bar{\mathcal{F}}_k(\boldsymbol{\theta}) ||^2  \bigg),
	\end{align}
	which is obtained after incorporating the expansion of $||\boldsymbol{\theta} - \boldsymbol{\theta}^\star - \eta \nabla \bar{\mathcal{F}}_k(\boldsymbol{\theta}) ||^2$. Substituting the GD update $\boldsymbol{\theta}' = \boldsymbol{\theta} - \eta\nabla\bar{\mathcal{F}}_k(\boldsymbol{\theta}) $ into (\ref{eq2}), we have 
	\begin{align}
	\bar{\mathcal{F}}_k(\boldsymbol{\theta}') - \bar{\mathcal{F}}_k(\boldsymbol{\theta}^\star) \leq \frac{1}{2\eta } \bigg(||\boldsymbol{\theta} - \boldsymbol{\theta}^\star   ||^2 - ||\boldsymbol{\theta}' - \boldsymbol{\theta}^\star   ||^2   \bigg).
	\end{align}
	Now, let us replace $\boldsymbol{\theta}'$ with $\boldsymbol{\theta}^{(i)}$, then summing over $i = 1,\dots, t$ yields
	\begin{align}
	&\sum_{i = 1}^{t} (\bar{\mathcal{F}}_k(\boldsymbol{\theta}^{(i)}) - \bar{\mathcal{F}}_k(\boldsymbol{\theta}^\star))  \nonumber \\
	& \leq \sum_{i = 1}^{t} \frac{1}{2\eta } \bigg(||\boldsymbol{\theta}^{(i-1)} - \boldsymbol{\theta}^\star   ||^2 - ||\boldsymbol{\theta}^{(i)} - \boldsymbol{\theta}^\star   ||^2   \bigg) \nonumber \\
	& = \frac{1}{2\eta }\bigg(||\boldsymbol{\theta}^{(0)} - \boldsymbol{\theta}^\star   ||^2 - ||\boldsymbol{\theta}^{(t)} - \boldsymbol{\theta}^\star   ||^2   \bigg) \nonumber \\
	& \leq  \frac{1}{2\eta }||\boldsymbol{\theta}^{(0)} - \boldsymbol{\theta}^\star  ||^2,   \label{eq4}
	\end{align}
	where the summation on the right hand side disappears since the consecutive terms cancel each other. Since $\bar{\mathcal{F}}_k(\boldsymbol{\theta}^{(t)})$ is a decreasing function, we have 
	\begin{align}
	\label{eq3}
	\bar{\mathcal{F}}_k(\boldsymbol{\theta}^{(t)}) - \bar{\mathcal{F}}_k(\boldsymbol{\theta}^\star) \leq  \frac{1}{t} \sum_{i=1}^t(\bar{\mathcal{F}}_k(\boldsymbol{\theta}^{(i)}) - \bar{\mathcal{F}}_k(\boldsymbol{\theta}^\star)).
	\end{align}
	Inserting (\ref{eq4}) into (\ref{eq3}), we finally have
	\begin{align}
	\bar{\mathcal{F}}_k(\boldsymbol{\theta}^{(t)}) - \bar{\mathcal{F}}_k(\boldsymbol{\theta}^\star) \leq \frac{1}{2\eta t}||\boldsymbol{\theta}^{(0)} - \boldsymbol{\theta}^\star  ||^2 .
	\end{align}

	\bibliographystyle{IEEEtran}
	\bibliography{IEEEabrv,references_085}
	\balance

\end{document}